\documentclass{article}
\usepackage{iclr2026_conference,times}

\iclrfinalcopy

\usepackage{listings}
\usepackage{adjustbox}
\usepackage{booktabs}
\usepackage{enumitem}
\RequirePackage{tgtermes}
\RequirePackage{newtxtext}
\RequirePackage{newtxmath}
\RequirePackage{bm}
\RequirePackage{amsmath}

\usepackage{multirow}
\usepackage{makecell}
\usepackage{array}
\usepackage{hyperref}
\usepackage{subcaption}
\usepackage{tcolorbox}
\tcbuselibrary{skins}
\usepackage{xcolor}
\usepackage{colortbl}
\usepackage{graphicx}

\usepackage{anyfontsize}
\usepackage[T1]{fontenc}
\usepackage{algorithm}
\usepackage{algpseudocode}
\usepackage{tikz}
\usetikzlibrary{
  arrows.meta,
  positioning,
  shapes,
  calc,
  fit,
  backgrounds
}
\usepackage{pgfplots}
\pgfplotsset{compat=1.18}

\definecolor{highlight}{rgb}{0.85, 0.95, 0.85}

\lstdefinestyle{minizinc}{
  basicstyle=\ttfamily\scriptsize,
  breaklines=true,
  frame=single,
  backgroundcolor=\color{gray!10}
}

\lstdefinelanguage{MiniZinc}{
  morekeywords={int, var, array, of, set, string, constraint, forall, sum, solve, minimize, in, div, array2d},
  sensitive=true,
  morecomment=[l]{\%},
  morestring=[b]",
}

\definecolor{bugbg}{HTML}{FFF5F5}
\definecolor{bugframe}{HTML}{E53E3E}
\definecolor{fixbg}{HTML}{F0FFF4}
\definecolor{fixframe}{HTML}{38A169}
\definecolor{codegray}{HTML}{F7F7F7}
\definecolor{kw}{HTML}{7B3294}
\definecolor{cm}{HTML}{808080}
\definecolor{st}{HTML}{2E7D32}
\definecolor{hlred}{HTML}{FED7D7}
\definecolor{hlgreen}{HTML}{C6F6D5}

\newcommand\ttm[1]{\textsc{Text2Model}}
\newcommand\ttz[1]{\textsc{Text2Zinc}}
\newcommand\ltz[1]{\textsc{Learn2Zinc-Base}}
\newcommand\mnz[1]{\textsc{MiniZinc}}

\newcommand{\best}[1]{\textbf{#1}}
\newcommand{\ptc}{\textsc{Learn2Zinc-Base}}
\newcommand{\ptcr}{\textsc{Learn2Zinc-CoT}}
\newcommand{\ptcpr}{\textsc{Learn2Zinc-Base+CoT}}

\usepackage{natbib}
\bibpunct[, ]{(}{)}{,}{a}{}{,}%

\title{\textsc{Learn2Zinc}: Fine-tuning Small Language Models for Text-to-Model Translation in \textsc{MiniZinc}}

\author{Serdar Kad{\i}o{\u{g}}lu\textsuperscript{1, 2} and Karthik Uppuluri\textsuperscript{1}\\
\textsuperscript{1}AI Center of Excellence, Fidelity Investments \\
\textsuperscript{2}Department of Computer Science, Brown University\\
\texttt{serdark@cs.brown.edu}}

\begin{document}

\maketitle

\begin{abstract}
Large language models excel at code generation for mainstream programming languages but struggle with rare, domain-specific languages such as \textsc{MiniZinc}, a constraint modeling language for combinatorial problems. We investigate whether targeted fine-tuning can teach small language models (0.6B to 20B parameters) to generate syntactically correct and semantically valid \textsc{MiniZinc} models from natural language problem descriptions. Our key finding is that syntax errors dominate failures when working with this domain specific language: the out-of-the-box execution accuracy of small language models such as Qwen3, LLaMa, Gemma, and GPT-OSS is near-zero. We propose a cross-model error bootstrapping approach that collects syntax errors from multiple LLM runs and leverage those to curate an error correction training dataset. 
This dataset allows us fine-tune small language models that consistently improves both direct code generation and chain-of-thought approaches across all model sizes. With self-reflection and ensembling, our approach achieves up to 98\% execution accuracy. In parallel, solution accuracy still remains at 35\%, indicating that while syntax is learnable, constraint reasoning remains a challenge. We contribute our fine-tuning pipeline, datasets, and models to opens-source for further research on text-to-model translation.
\end{abstract}

\vspace{-0.2cm}
\section{Introduction}
\label{sec:intro}

Optimization technology has achieved significant advancements, ranging from dramatic improvements in solver efficiency to the development of high-level modeling languages to enhance usability. Nevertheless, the fundamental decision-making framework has remained unchanged for decades, adhering to the de facto \textit{model-and-run strategy}. Within this status quo, users are required to manually convert problem descriptions into optimization models, which are subsequently processed by solvers to obtain solutions.

Over the years, high-level modeling languages such as \textsc{MiniZinc}~\citep{minizinc}, CPMpy~\citep{guns2019increasing}, and GAMS~\citep{Bussieck2004} have partially addressed the accessibility challenge by providing solver-agnostic approaches that are powerful and flexible. These modeling frameworks enable practitioners to focus on describing their problems without worrying about specific solution methods, making them especially useful for real-world applications, where requirements often change over time. However, the cognitive barrier of translating problem descriptions into formal constraint models persists. This barrier is particularly acute, as domain experts who deeply understand their problem domain often lack the specialized knowledge required for formal modeling. The resulting dependency on modeling experts creates operational bottlenecks and can lead to misinterpretation of domain-specific requirements during the translation process.

In parallel, Large Language Models (LLMs) have emerged as the communication medium with machines~\citep{openai2024gpt4technicalreport,geminiteam2025geminifamilyhighlycapable,deepseekai2025deepseekv3technicalreport}. While language models are powerful at interfacing with natural text, they struggle with the consistency and precision required in formal, declarative approaches, from basic type declarations to complex constraint relations. They face significant challenges in handling the mathematical and logical reasoning needed for text-to-model translation~\citep{SimchiLevi2025Democratizing,Wasserkrug2025,ner4opt2024}\footnote{\url{https://skadio.github.io/text2model}}. 

This current gap between understanding textual descriptions and turning them into problem formulations indicates that more work is needed for modeling assistants.


A critical observation motivating this work is that \textsc{MiniZinc}, as a domain-specific language, has far less representation in typical LLM pretraining corpora compared to established languages such as Python, C++, or Java. For perspective, the GitHub topic page for \textsc{MiniZinc} lists on the order of a hundred repositories, compared to millions for Python or JavaScript. This limited presence means that LLMs have had minimal exposure to \textsc{MiniZinc} syntax during their pretraining. Supporting this observation, four out of five models we tested; Qwen3, LLaMa, and Gemini, achieve 0.0\% execution accuracy on \textsc{MiniZinc} generation, and the largest model (GPT-OSS-20B) reaches only 6.0\%. This confirms that \textsc{MiniZinc} is clearly out-of-distribution for the current best small language models. This raises a fundamental research question: \textit{Can targeted fine-tuning teach small language models a domain-specific programming language such as \textsc{MiniZinc}?}. This is exactly what we study in this paper.

\subsection{Our Contributions}

Our contributions are as follows:

\begin{enumerate}
    \item We present the first systematic study of fine-tuning small language models from 0.6B to 20B parameters for \textsc{MiniZinc} code generation.
    
   \item To make fine-tuning possible, we introduce a cross-model error bootstrapping approach that collects syntax errors from multiple LLM runs and leverage this to create a realistic error correction training dataset.
    
    \item We propose three fine-tuning strategies of increasing sophistication and show that augmenting the training data with error correction examples consistently outperforms both direct code generation and chain-of-thought approaches across all model sizes.
    
    \item Our performance, when using an ensemble of our fine-tuned models, \textbf{achieves 98\% execution accuracy, up from 0.0-6.0\% out-of-the-box accuracy, effectively solving the \textsc{MiniZinc} syntax problem}. At the same time, we identify constraint reasoning as the remaining bottleneck with only 35\% solution accuracy with a detailed analysis on error modes.
    
    \item We open-source our fine-tuning pipeline \footnote{\url{https://github.com/skadio/learn2zinc}}, the best-performing fine-tuned model at various scale 0.6B, 1B, 3B, 9B, and 20B parameters\footnote{\url{https://huggingface.co/skadio/learn2zinc}}, and our instruction-tuning datasets\footnote{\url{https://huggingface.co/datasets/skadio/learn2zinc}} to facilitate reproducibility and future research in text-to-model generation.
\end{enumerate}

\section{Background}
\label{sec:background}

Let us briefly review the constraint modeling language used in this study, \textsc{MiniZinc}, the dataset used for benchmarking, \textsc{Text2Zinc}, and our LLM copilot approaches, \textsc{Text2Model}.
 
\subsection{\textsc{MiniZinc}}
 
\textsc{MiniZinc}~\citep{minizinc} is a high-level constraint modeling language that supports both discrete and continuous optimization and satisfaction problems. Its solver-agnostic design allows communication with various solver backends, including Constraint Programming (CP), (Mixed) Integer Programming (MIP), and Boolean Satisfiability/Lazy Clause Generation (SAT). This flexibility is achieved through compilation to \textsc{FlatZinc}, an intermediate language that interfaces with different solvers, allowing the same \textsc{MiniZinc} model to be used across multiple backends without code modifications.
 
A key feature of \textsc{MiniZinc} is its use of global constraints, which significantly simplify the modeling process. For example, the \texttt{all\_different} constraint specifies that a set of variables must take distinct values, replacing numerous pairwise inequality constraints. 

The \textsc{MiniZinc} language structure consists of four main components: decision variables, constraints, parameters, and an objective function (for optimization) or a satisfaction goal. \textsc{MiniZinc} also separates models (\texttt{.mzn} files) from data instances (\texttt{.dzn} files), allowing a single model to be reused across multiple problem instances.

This paper builds on two previous complementary works:  \textsc{Text2Zinc} dataset to establish a common benchmark for this task, and \textsc{Text2Model} copilots to establish a baseline performance of various LLM approaches. 
 
\subsection{\textsc{Text2Zinc} Dataset}
\textsc{Text2Zinc}~\citep{text2zinc} introduces a cross-domain dataset for modeling optimization and satisfaction problems in \textsc{MiniZinc}. \textit{It is the first dataset in this line of research that covers both satisfaction and optimization problems in a solver- and paradigm-agnostic language.} The dataset brings together 1,775 problems from multiple sources, including \textsc{Nlp4lp}~\citep{optimus}, \textsc{Hakank}, \textsc{ComplexOr}~\citep{complexor}, \textsc{LpWp}~\citep{lpwp}, \textsc{CspLib}, and problems from Cardinal Operations covering both \textsc{Mamo} and \textsc{Nl4Opt}~\citep{ner4opt2024} collections (see Table~\ref{tab:problem_sources} for a full breakdown by category). Of these, 110 are fully verified with manually written \textsc{MiniZinc} models, complete metadata, and validated solutions. For the remaining problems, which originate from \textsc{IndustryOr}, \textsc{Mamo}, and \textsc{Nl4Opt}, ground-truth objective values are available through the original sources for verification purposes. The dataset provides $is\_optimization$, $is\_satisfaction$, $has\_verified\_obj$, $has\_verified\_mzn$, $has\_dzn$ to distinguish among these properties.

\subsection{\textsc{Text2Model} Copilots}
\textsc{Text2Model}~\citep{text2model} introduces a suite of copilots for text-to-model translation using frontier LLMs. It evaluates several strategies of varying complexity on the \textsc{Text2Zinc} dataset, including zero-shot prompting, chain-of-thought reasoning, knowledge-graph representations, grammar-based syntax encoding, and agentic approaches. A key observation is that even frontier LLMs with sophisticated prompting strategies are not \textit{yet} a push-button technology for combinatorial modeling.

\section{Learn2Zinc: Fine-Tuning Small Language Models}
\label{sec:learn2zinc}
The important insight that led to this work is as follows. Running several \textsc{Text2Model} copilots on hundreds of \textsc{Text2Zinc} instances generate a considerable amount \textsc{MiniZinc} models,  even though the \textsc{MiniZinc} models are not manually written. While generation is straightforward, the key advantage stems from the \textit{verification loop}: the output can be verified with known objective value for optimization problems, and feasibility can be asserted for satisfaction problems. This yields a large pool of verified $\langle text, model \rangle$ pairs that makes fine-tuning possible. 
 
In this paper, we go beyond dependency on large frontier models for generating constraint models and investigate  whether targeted fine-tuning can teach \textit{small language models} (0.6B--20B parameters) to generate \textsc{MiniZinc} code. Crucially, this effort requires designing a fine-tuning dataset. Our fine-tuning dataset starts from the verified \textsc{MiniZinc} solutions in \textsc{Text2Zinc}. In addition, we draw samples leveraging the \textsc{Or-Instruct} dataset~\citep{orlm}\footnote{We are indebted to the creators of the \textsc{Or-Instruct} dataset for their valuable contribution to the community.}. The \textsc{Or-Instruct} dataset contains optimization problems with \textsc{Copt} models written in Python. Given these problem descriptions, their \textsc{Copt} model, and verified objective value, we use GPT-5.2 to generate a translation from \textsc{Copt} to \textsc{MiniZinc}. We add the result to our fine-tuning dataset if and only if \textsc{MiniZinc} output matches the known objective. 

As shown in Table~\ref{tab:problem_sources}, when \textsc{MiniZinc} models for \textsc{Text2Zinc} and \textsc{Or-Instruct} are combined, they together cover a set of \textbf{2,208} unique problems. Importantly, instances might be associated with alternative \textsc{MiniZinc} models generated by different \textsc{Text2Model} copilot strategy. Overall, this yields \textbf{8,014} instruction-tuning $\langle text, model \rangle$ pairs for fine-tuning small language models.

\begin{table}[t]
\centering
\begin{tabular}{llrr}
\toprule
Dataset & Source & Instances & Percentage \\
\midrule
\multirow{1}{*}{\textsc{Orlm}~\citep{orlm}} 
& \textsc{Or-Instruct-3K} & 1,351 & 61.2\% \\
\cmidrule{1-4} 
\multirow{7}{*}{\textsc{Text2Zinc}~\citep{text2zinc}}
& \textsc{Mamo} & 604 & 27.4\% \\
& \textsc{Nl4Opt} & 200 & 9.1\% \\
& \textsc{Nlp4Lp} & 30 & 1.4\% \\
& \textsc{Hakank} & 10 & 0.5\% \\
& \textsc{ComplexOr} & 5 & 0.2\% \\
& \textsc{LpWp} & 5 & 0.2\% \\
& \textsc{CspLib} & 3 & 0.1\% \\
\midrule
& \textbf{Total} & \cellcolor{highlight}\textbf{2,208} & \cellcolor{highlight}\textbf{100\%} \\
\bottomrule
\end{tabular}
\caption{Distribution of unique problem instances by source. Problems from \textsc{Text2Zinc} were verified against known objectives using \textsc{Text2Model} copilot strategies. \textsc{Or-Instruct} problems were translated from \textsc{Copt} models written in Python to \textsc{MiniZinc} via GPT-5.2.}
\label{tab:problem_sources}
\end{table}

\section{Learn2Zinc: Fine-tuning Methodology}
\label{sec:finetuning}

Our fine-tuning methodology consists of a combination of different small language models at varying parameter sizes and different fine-tuning datasets. We consider both a \textit{generation fine-tuning} and \textit{error-correction fine-tuning} to teach SLMs \textsc{MiniZinc} syntax (\textsection ~\ref{sec:teaching}). 

\subsection{Small Language Models (Slms)}

We consider four different families of small languages models (SLMs): Qwen, LLaMa, Gemma, and GPT. In particular, we fine-tune five different variants to cover different parameter sizes: Qwen3-0.6B (the smallest to test the lower bound of capability), LLaMA-3.2-1B and LLaMA-3.2-3B, Gemma-2-9B, and GPT-OSS-20B (the largest number of parameters).

\subsection{Datasets \& Setup}

We consider three different fine-tuning instruction datasets to consider a baseline, a chain-of-thought, and a mixed strategy. 

\paragraph{\textsc{Learn2Zinc-Base}.} We obtain 8,014 instances with $\langle text, model \rangle$ pairs as described in \textsection ~\ref{sec:finetuning}.


\paragraph{\textsc{Learn2Zinc-CoT} (Chain-of-Thought).} We take the baseline dataset with $\langle text, model \rangle$ pairs, and prompt a reasoning model, GPT4o. to generate reasoning chain from the problem text to the constraint model and identify variables, parameters, constraints, objective. Accordingly, this dataset also has 8,014 pairs of the form $\langle text, reasoning, model \rangle$.


\paragraph{\textsc{Learn2Zinc-Base+CoT}.} We combine both \textsc{Learn2Zinc-Base} and \textsc{Learn2Zinc-CoT} examples leading to 16,028 pairs.

In Appendix~\ref{appendix:examples}, we present examples of our baseline and CoT pairs. Notice how the CoT version introduces the reasoning chain as an intermediary in between the problem description and the constraint model.


In our experiments, we fine-tune five different SLMs on three different fine-tuning datasets. The details of training hyperparameters are given in Appendix-Table~\ref{tab:training_config}. We employ Low-Rank Adaptation (LoRA) with 8-bit quantization for all models, except 4-bit for GPT-OSS-20B, to fit on a single A100 GPU. 

We choose LoRA over full fine-tuning for two reasons. First, our fine-tuning datasets are relatively small, with the base dataset comprising only 8,014 instances. At this scale, full fine-tuning is prone to overfitting and offers limited gains over parameter-efficient alternatives. Second, LoRA significantly reduces memory and compute requirements, allowing us to fine-tune models ranging from 0.6B to 20B parameters on a single GPU, where full fine-tuning would not be feasible. Among parameter-efficient fine-tuning (PEFT) approaches, LoRA is well-established and widely adopted, offering a strong balance between task adaptation and training efficiency, making it a suitable choice for our experimental setting.

\begin{table}[t]
\centering
\begin{tabular}{llcc}
\toprule
\textbf{Model} & \textbf{Strategy} & \textbf{Exec Acc (\%)} & \textbf{Sol Acc (\%)} \\
\midrule
\multirow{5}{*}{Qwen3-0.6B} 
& Original-8bit & 0.0 & 0.0 \\
& \textsc{Learn2Zinc-Base} & 51.0 & 9.0 \\
& \textsc{Learn2Zinc-CoT} & 44.0 & 10.0 \\
& \textsc{Learn2Zinc-Base+CoT} & 51.0 & 12.0 \\
& \textsc{Learn2Zinc-Augmented} & \cellcolor{highlight}\best{64.0} & \cellcolor{highlight}\best{13.0} \\
\midrule
\multirow{5}{*}{LLaMA-3.2-1B} 
& Original-8bit & 0.0 & 0.0 \\
& \textsc{Learn2Zinc-Base} & 44.0 & 4.0 \\
& \textsc{Learn2Zinc-CoT} & 22.0 & 1.0 \\
& \textsc{Learn2Zinc-Base+CoT} & 46.0 & 4.0 \\
& \textsc{Learn2Zinc-Augmented} & \cellcolor{highlight}\best{57.0} & \cellcolor{highlight}\best{8.0} \\
\midrule
\multirow{5}{*}{LLaMA-3.2-3B}
& Original-8bit & 0.0 & 0.0 \\
& \textsc{Learn2Zinc-Base} & 62.0 & 10.0 \\
& \textsc{Learn2Zinc-CoT} & 47.0 & 9.0 \\
& \textsc{Learn2Zinc-Base+CoT} & 58.0 & 12.0 \\
& \textsc{Learn2Zinc-Augmented} & \cellcolor{highlight}\best{70.0} & \cellcolor{highlight}\best{17.0} \\
\midrule
\multirow{5}{*}{Gemma-2-9B} 
& Original-8bit & 0.0 & 0.0 \\
& \textsc{Learn2Zinc-Base} & \cellcolor{highlight}\best{74.0} & \cellcolor{highlight}\best{22.0} \\
& \textsc{Learn2Zinc-CoT} & 49.0 & 15.0 \\
& \textsc{Learn2Zinc-Base+CoT} & 70.0 & 21.0 \\
& \textsc{Learn2Zinc-Augmented} & 72.0 & \cellcolor{highlight}\best{22.0} \\
\midrule
\multirow{5}{*}{GPT-OSS-20B} 
& Original-4bit & 6.0 & 5.0 \\
& \textsc{Learn2Zinc-Base} & 66.0 & 27.0 \\
& \textsc{Learn2Zinc-CoT} & 52.0 & 21.0 \\
& \textsc{Learn2Zinc-Base+CoT} & 63.0 & 27.0 \\
& \textsc{Learn2Zinc-Augmented} & \cellcolor{highlight}\best{76.0} & \cellcolor{highlight}\best{32.0} \\
\midrule
GPT-5.2\textsuperscript{\dag} & \textsc{Agentic + Code} (\cite{text2model}) & 86.0 & \best{57.0} \\
\bottomrule
\end{tabular}
\caption{Execution and Solution Accuracy (\%) across models and fine-tuning strategies. Best results per model are highlighted. \textsuperscript{\dag}GPT-5.2 results are from \cite{text2model} and shown for reference.}
\label{tab:main_results}
\end{table}

\subsection{Evaluation Metrics}
We use Execution Accuracy (the constraint model compiles and runs) and Solution Accuracy (the objective value found by solving the constraint model matches the ground truth) as used previously in several studies, e.g., ~\cite{text2model}. 

\subsection{Numerical Results}
For benchmarking, we use \textsc{Text2Zinc-IndustryOr} (100 problems) as our test set. It is important to note theses problems are\textit{ not included in the fine-tuning}. We use \textsc{IndustryOr} since it is the most challenging benchmark, as reported in~\cite{text2model,orlm,lean-llm-opt}, which shows 86\% execution accuracy and 57\% solution accuracy with an Agentic copilot using the frontier GPT-5.2 model. As the underlying solver, we use HiGHS in our experiments, with a 120-second timeout per problem.

\begin{figure}[t]
\centering
\begin{tikzpicture}
\begin{axis}[
    width=\textwidth,
    height=6.5cm,
    ybar,
    bar width=4pt,
    ylabel={Execution Accuracy (\%)},
    symbolic x coords={Qwen3-0.6B, LLaMA-1B, LLaMA-3B, Gemma-9B, GPT-OSS-20B},
    xtick=data,
    xticklabel style={font=\small},
    ymin=0, ymax=100,
    legend style={
        at={(0.5,1.02)},
        anchor=south,
        legend columns=4,
        font=\small,
        /tikz/every even column/.append style={column sep=6pt}
    },
    grid=major,
    grid style={gray!30},
    ymajorgrids=true,
    xmajorgrids=false,
    enlarge x limits=0.15,
]
\addplot[fill=gray!50, draw=gray!70] coordinates {
    (Qwen3-0.6B, 51) (LLaMA-1B, 44) (LLaMA-3B, 62) (Gemma-9B, 74) (GPT-OSS-20B, 66)
};
\addplot[fill=blue!40, draw=blue!60] coordinates {
    (Qwen3-0.6B, 44) (LLaMA-1B, 22) (LLaMA-3B, 47) (Gemma-9B, 49) (GPT-OSS-20B, 52)
};
\addplot[fill=orange!40, draw=orange!60] coordinates {
    (Qwen3-0.6B, 51) (LLaMA-1B, 46) (LLaMA-3B, 58) (Gemma-9B, 70) (GPT-OSS-20B, 63)
};
\addplot[fill=green!50, draw=green!70] coordinates {
    (Qwen3-0.6B, 64) (LLaMA-1B, 57) (LLaMA-3B, 70) (Gemma-9B, 72) (GPT-OSS-20B, 76)
};
\legend{\textsc{Learn2Zinc-Base}, \textsc{Learn2Zinc-CoT}, \textsc{Learn2Zinc-Base+CoT}, \textsc{Learn2Zinc-Augmented}}
\end{axis}
\end{tikzpicture}
\caption{Execution accuracy (\%) across fine-tuning strategies and model sizes. \textsc{Learn2Zinc-Augmented} (green) consistently achieves the highest execution accuracy across all models, while \textsc{Learn2Zinc-CoT} (blue) underperforms all other strategies.}
\label{fig:exec_accuracy}
\end{figure}
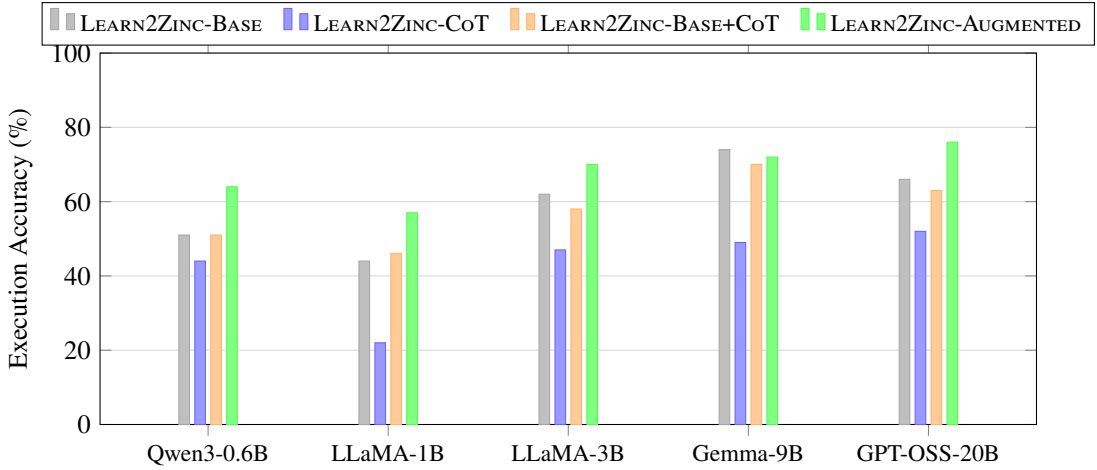

Table~\ref{tab:main_results} presents execution and solution accuracy for each model and fine-tuning strategy. Please omit the augmented row in the table for now as we discuss this later in \textsection ~\ref{sec:teaching}. Figure~\ref{fig:exec_accuracy} visualizes the same results to reveal emerging patterns. 

Several patterns emerge from results in Table~\ref{tab:main_results}. First of all, \textbf{original SLMs achieve near-zero accuracy}. All models score 0\%, except for GPT-OSS-20B with 6.0\% execution and 5.0\% solution accuracy. This confirms the main motivation of this paper: \textit{\textsc{MiniZinc} is clearly out-of-distribution for current best SLMs}. \textbf{With our fine-tuning, Gemma-2-9B achieves a remarkable boost from its original 0\% execution accuracy to 74\%}. As a comparison between SLMs and frontier models, according to~\cite{text2model}, \textbf{GPT-5.2 performs 86\% execution and 57\% solution accurac on the same test set} to generate \textsc{MiniZinc} models, as reported at the bottom row of Table~\ref{tab:main_results}. 

As Figure~\ref{fig:exec_accuracy} makes it clear, interestingly, \textsc{Learn2Zinc-CoT} underperforms \textsc{Learn2Zinc-Base} across all  models. This finding contrasts with several existing results where chain-of-thought often helps~\cite{text2model,text2zinc}, hinting that, unlike frontier models, SLMs cannot take advantage of reasoning traces. A possible interpretation is, reasoning traces cannot compensate for missing syntax when the small model has not been trained for \textsc{MiniZinc} at scale. Similarly, simply mixing \textsc{Learn2Zinc-Base} and \textsc{Learn2Zinc-CoT} examples provides no consistent improvement over \textsc{Learn2Zinc-Base} alone. Based on these findings, we therefore pursue a dedicated approach next to teach SLMs \textsc{MiniZinc} syntax.

\section{Teaching Slms \textsc{MiniZinc} Syntax}
\label{sec:teaching}
Our initial fine-tuning experiments with five SLMs and three fine-tuning strategies reveals a consistent pattern: \textit{SLM outputs are riddled with syntax errors}. SLMs, even with our careful fine-tuning, cannot produce correct \textsc{MiniZinc} code where the best execution accuracy is still at 76\%. Before we could even evaluate whether SLMs capture correct combinatorial semantics, most outputs fail to capture \textsc{MiniZinc} syntax and compile. Execution accuracy, even before solution accuracy, remains an immediate bottleneck.

This observation shifts our focus. If we could potentially improve execution rates, models would have more opportunity to produce correct solutions. The question then becomes: \textit{how can we teach SLMs to avoid and correct their syntax mistakes?} So far, we approached fined-tuning for text-to-model translation as \textit{generation fine-tuning} and we now shift our focus to \textit{error-correction fine-tuning}. 

For this purpose, we design an augmented training strategy that \textit{explicitly teaches syntax correction}. The key insight is that models need exposure to common errors \textit{and their fixes}, not just correct outputs.

\begin{table}
\hspace{-1.5cm}
\renewcommand{\arraystretch}{1.3}
\begin{tabular}[t]{llp{5cm}p{4cm}}
\toprule
Rule ID & Difficulty & Description & Example \\
\midrule
\texttt{E1\_drop\_semicolon} & Easy & Remove trailing semicolon from a statement & var int: x = 5;  $\rightarrow$  var int: x = 5 \\
\texttt{E2\_drop\_comma} & Easy & Remove a comma from a list, array, or parameter list & [1, 2, 3]  $\rightarrow$  [1 2, 3] \\
\texttt{E3\_2d\_array\_open} & Easy & Replace [| with [ in 2D array literal & [| 1, 2 | 3, 4 |]  $\rightarrow$  [ 1, 2 | 3, 4 |] \\
\texttt{E4\_2d\_array\_close} & Easy & Replace |] with ] in 2D array literal & [| 1, 2 | 3, 4 |]  $\rightarrow$  [| 1, 2 | 3, 4 ] \\
\texttt{E5\_drop\_row\_pipe} & Easy & Remove | row separator in 2D array & [| 1, 2 | 3, 4 |]  $\rightarrow$  [| 1, 2   3, 4 |] \\
\texttt{E6\_drop\_close\_paren} & Easy & Remove a closing parenthesis & sum(x)  $\rightarrow$  sum(x \\
\texttt{E7\_drop\_close\_bracket} & Easy & Remove a closing bracket & array[1..N]  $\rightarrow$  array[1..N \\
\texttt{M1\_drop\_var} & Medium & Remove "var" keyword from variable declaration & var int: x  $\rightarrow$  int: x \\
\texttt{M2\_drop\_of} & Medium & Remove "of" keyword from type expression & set of int  $\rightarrow$  set int \\
\texttt{M3\_solve\_keyword} & Medium & Replace "maximize"/"minimize" with invalid "max"/"min" & solve maximize obj  $\rightarrow$  solve max obj \\
\texttt{M4\_capitalize\_keyword} & Medium & Capitalize a \textsc{MiniZinc} keyword (case-sensitive error) & constraint x > 0  $\rightarrow$  Constraint x > 0 \\
\texttt{M5\_split\_endif} & Medium & Split "endif" into "end if" or "elseif" into "else if" & endif  $\rightarrow$  end if \\
\texttt{M6\_drop\_gen\_in} & Medium & Remove "in" from generator expression & forall(i in 1..N)  $\rightarrow$  forall(i 1..N) \\
\texttt{H1\_drop\_then} & Hard & Remove "then" from if-then-else expression & if x > 0 then y  $\rightarrow$  if x > 0 y \\
\texttt{H2\_drop\_let\_in} & Hard & Remove "in" keyword after let block closing brace & let \{ var int: t \} in t  $\rightarrow$  let \{ var int: t \} t \\
\texttt{H3\_swap\_type\_ident} & Hard & Swap type and identifier in declaration & var int: x  $\rightarrow$  x: var int \\
\texttt{H4\_triple\_dot\_range} & Hard & Replace ".." range operator with "..." (invalid) & 1..N  $\rightarrow$  1...N \\
\texttt{H5\_wrong\_logic\_op} & Hard & Replace \texttt{/\textbackslash} with \texttt{\&\&} or \texttt{\textbackslash/} with \texttt{||} (invalid C-style operators) & \texttt{x /\textbackslash{} y}  $\rightarrow$  \texttt{x \&\& y} \\
\texttt{H6\_extra\_semicolon} & Hard & Insert extra semicolon in the middle of a constraint & constraint x + y = z  $\rightarrow$  constraint x +; y = z \\
\texttt{H7\_drop\_close\_brace} & Hard & Remove a closing brace \} & \{1, 2, 3\}  $\rightarrow$  \{1, 2, 3 \\
\bottomrule
\end{tabular}
\caption{Complete taxonomy of \textsc{MiniZinc} syntax corruption rules derived from the BNF grammar.}
\label{tab:corruption_taxonomy}
\end{table}

\subsection{Error Correction Dataset}
For error correction fine-tuning, we draw data samples from two different strategies; synthetic corruption and cross-model error bootstrapping. 

\paragraph{Synthetic Corruptions.} 
We leverage the Backus-Naur Form (BNF) grammar of \textsc{MiniZinc}\footnote{We thank Guido Tack for sharing this grammar with us.} to derive 20 corruption rules, each grounded in a specific grammar production. These rules are organized into three difficulty levels: easy (7 rules targeting delimiters and punctuation, e.g., missing semicolons, dropped commas, unbalanced brackets), medium (6 rules targeting keywords and types, e.g., invalid solver directives, dropped \texttt{var} or \texttt{of} keywords, split compound tokens like \texttt{endif}), and hard (7 rules targeting structural elements, e.g., swapped declaration order, invalid operators, misplaced semicolons). 

\begin{figure}[t!]
\centering
\includegraphics[width=\textwidth]{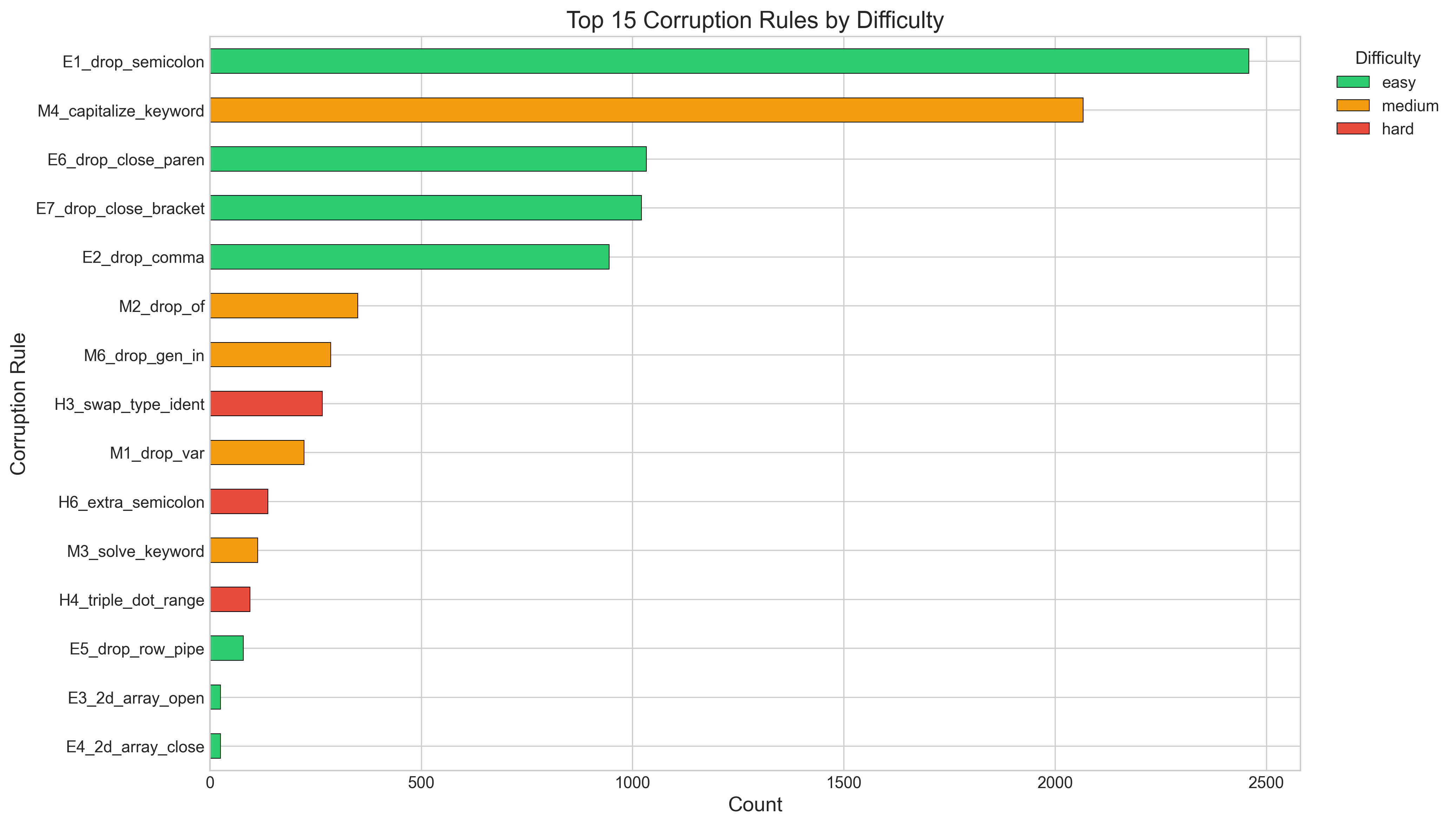}
\caption{Top-15 \textsc{MiniZinc} grammar corruption rules color coded by difficulty.}
\label{fig:rules_by_difficulty}
\end{figure}

Table~\ref{tab:corruption_taxonomy} lists the complete taxonomy of grammar-based corruption rules. To generate the training data, we sample instances from the baseline dataset and apply corruptions at  controlled difficulty with 30\% easy, 30\% medium, 30\% hard, and 10\% identity. Within each difficulty we sample the rule to apply proportionally with our empirical observations. In case of identity, the model needs to recognize that no fix is necessary. Each corrupted model is verified using the \textsc{MiniZinc} compiler to ensure the gold code passes and the corrupted code fails syntax checking; pairs that do not satisfy both conditions are discarded. This process yields \textbf{4,452 verified instances} with $\langle text, model, corrupt\_model \rangle$ triples. Figure~\ref{fig:rules_by_difficulty} shows the distribution of rules applied in the resulting dataset color coded by their difficulty.

\paragraph{Cross-Model Error Bootstrapping.} Our probabilistic synthetic error generation from \textsc{MiniZinc} grammar may not capture actual failure modes of LLMs at test time. We address this by collecting errors and their fixed version as follows: 

\begin{enumerate}
    \item We run all five SLMs across all three fine-tuning strategies (base, CoT, mixed) on the base dataset with multiple sampling temperatures to increase variety of error cases.
    \item When we obtain an execution failure, for each $\langle text, corrupt\_model, error\_message \rangle$, we employ GPT-5.2 to attempt a fix. A natural alternative would be to pair the failed output directly with a correct generation from the same model. However, we found that correct and incorrect generations from the same SLM often differ substantially in structure and variable naming, meaning the pairing would not represent a targeted fix but rather a complete rewrite. Such examples would poorly represent the error correction task. Instead, we prompt GPT-5.2 with the corrupted model and its error message, instructing it to make only the minimal changes necessary to fix the error while preserving the original code structure. This produces training pairs where the correction is localized to the actual fault.
    \begin{enumerate}
        \item If the fix succeeds, i.e, the updated model compiles \textit{and} produces the correct output, we add  $\langle text, corrupt\_model, correct\_model \rangle$ to our fine-tuning dataset. This process yields \textbf{2,286} verified instances and is aimed at teaching SLMS error correction. 
        
        \smallskip
        Table~\ref{tab:correction_distribution} shows the distribution of error-correction examples over SLMs varied by temperature. In total, the cross-model error bootstrapping process creates a diverse error corpus.  As expected, smaller models (Qwen3-0.6B and LLaMA-3.2-1B) contribute the most correction examples. Fine-tuning SLMs with this dataset has an interesting property: small models (e.g., Qwen3-0.6B) learn from large model (e.g., GPT-OSS-20B) mistakes, and vice versa. Higher sampling temperatures also produce more errors with temperature 0.8 accounting for 38.1\% of corrections vs. 29.4\% at temperature 0.2. 

        \smallskip   
        Table~\ref{tab:strategy_errors} shows the distribution of errors in this cross-model dataset by the fine-tuning strategy. Notice that, \textsc{Learn2Zinc-CoT} training produces the most errors during bootstrapping (43.9\%), suggesting again that reasoning traces without syntax knowledge may actually introduce more mistakes.

    \end{enumerate}
    
    \item Alternatively, when we obtain execution \textit{and} solution success, we collect a new positive generation example,  $\langle text, correct\_model \rangle$ to our fine-tuning dataset. This process yields \textbf{8,911 verified instances}, i.e., slightly extends our base dataset from 8,014, and is aimed at teaching SLMs constraint model generation. 
\end{enumerate}

\begin{table}[t]
\centering
\begin{tabular}{lrrr|rr}
\toprule
Model & T=0.2 & T=0.5 & T=0.8 & Total & \% \\
\midrule
LLaMA-3.2-1B & 230 & 262 & 286 & 778 & 34.0 \\
Qwen3-0.6B & 217 & 256 & 274 & 747 & 32.7 \\
LLaMA-3.2-3B & 139 & 126 & 192 & 457 & 20.0 \\
Gemma-2-9B & 85 & 101 & 118 & 304 & 13.3 \\
\midrule
Total & 671 & 745 & 870 & \cellcolor{highlight}\textbf{2,286} & \cellcolor{highlight}\textbf{100.0} \\
\% & 29.4 & 32.6 & 38.1 & & \\
\bottomrule
\end{tabular}
\caption{Distribution of correction pairs by error-producing SLM and sampling temperatures.}
\label{tab:correction_distribution}
\end{table}

\begin{table}[t]
\centering
\begin{tabular}{lrr}
\toprule
Strategy & Errors & Percentage \\
\midrule
\ptc{} & 640 & 28.0\% \\
\ptcr{} & 1,003 & 43.9\% \\
\ptcpr{} & 643 & 28.1\% \\
\midrule
Total & \cellcolor{highlight}\textbf{2,286} & \cellcolor{highlight}\textbf{100\%} \\
\bottomrule
\end{tabular}
\caption{Errors by training strategy during bootstrapping.}
\label{tab:strategy_errors}
\vspace{-0.2cm}
\end{table}

\begin{table}[b]
\centering
\begin{tabular}{ll}
\toprule
Task Type & Count \\
\midrule
Error-Correction tasks (total) & 6,738 \\
\quad - Synthetic corruptions & \quad 4,452 \\
\quad - Cross-model corrections & \quad 2,286 \\
Generation task & 8,911 \\
\midrule
\textbf{Total} & \cellcolor{highlight}\textbf{15,649} \\
\bottomrule
\end{tabular}
\caption{Breakdown of the augmented dataset by task type.}
\label{tab:model_corrections}
\end{table}

Overall, as shown in Table~\ref{tab:model_corrections}, when we combine cross-model error correction bootstrapping instances (6,728) and generation task instances together (8,911), we obtain \textbf{15,649 instances} for fine-tuning.

\subsection{\textsc{Learn2Zinc-Augmented} Fine-Tuning}

Given the the augmented fine-tuning dataset in Table~\ref{tab:model_corrections}, which hosts 15,649 instances in combination, we conduct the same fine-tuning protocol on all five SLMs. This produces our \textsc{Learn2Zinc-Augmented} fine-tuned model families. Notice that compared to our previous fine-tuning in \textsection \ref{sec:finetuning}, this augmented fine-tuning has two simultaneous objectives; a generation task and an error-correction task.

As previously reported in Table~\ref{tab:main_results}, the \textsc{Learn2Zinc-Augmented} strategy achieves the best execution accuracy on all models, except one, and more importantly, achieves the best solution accuracy among all approaches. When analyzing SLMs individually, Qwen3-0.6B jumps from 0.0\% execution to 51\% with our initial fine-tuning, and then to 65\% accuracy with our augmented fine-tuning. Similarly, LLaMa-3.2-1B goes from 0.0\%  to 46\% and then to 57\%. Gemma-2-9B goes from 0.0\%  to 74\% and 72\%. GPT-OSS-20B goes from 6.0\% to 66\% and then to 76\%. \textbf{Our best zero-shot results achieves 76\% using augmented GPT-OSS-20B}. This validates that \textit{training for error correction directly addresses the syntax bottleneck.}

\section{\textsc{Learn2Zinc} with Self-Reflection \& Ensemble}
So far, we only considered a zero-shot strategy, even though we fined-tuned for error correction.  As such, we next utilize a \textit{self-reflection loop} and \textit{ensemble} on top of the \textsc{Learn2Zinc-Augmented} strategies. Since other fine-tuning strategies have never seen error correction examples, we only consider results of \textsc{Learn2Zinc-Augmented} strategy for self-reflection and ensembling. 

\subsection{Self-Reflection Loop}
We run augmented fine-tuned models up to five attempts to produce executable code. On each failed attempt, the model receives its broken code and the compiler error message, then tries again, until hitting the limit.

As shown in Table~\ref{tab:retry}, adding reflection improves execution accuracy considerably and slightly improves solution accuracy.
Notice that reflections reach a running model, on average, \textit{in less than two trials}, with larger SLMs requiring fewer attempts. 

\begin{table}[t]
\centering
\renewcommand{\arraystretch}{1.25}
\begin{tabular}{l|cc|ccc}
\toprule
& \multicolumn{2}{c|}{\textbf{Single Pass}} & \multicolumn{3}{c}{\textbf{Self-Reflection@5}} \\
\textbf{Model} & \textbf{Exec (\%)} & \textbf{Sol (\%)} & \textbf{Exec (\%)} & \textbf{Sol (\%)} & \textbf{Avg Attempts} \\
\midrule
Qwen3-0.6B-Augmented & 64.0 & 13.0 & 72.0 & 13.0 & 2.22 \\
LLaMA-3.2-1B-Augmented & 57.0 & 8.0 & 71.0 & 8.0 & 2.35 \\
LLaMA-3.2-3B-Augmented & 70.0 & 17.0 & 80.0 & 17.0 & 1.90 \\
Gemma-2-9B-Augmented & 72.0 & 22.0 & 84.0 & 24.0 & 1.79 \\
GPT-OSS-20B-Augmented & 76.0 & 32.0 & \cellcolor{highlight} \best{89.0} & \cellcolor{highlight} \best{34.0} & \cellcolor{highlight} \best{1.62} \\
\bottomrule
\end{tabular}
\caption{\textsc{Learn2Zinc-Augmented Slms}: Single-pass vs. Self-reflection with five attempts.}
\label{tab:retry}
\end{table}

Execution accuracy improves substantially with retries, confirming that \textsc{Learn2Zinc-Augmented} training teaches models to fix their own mistakes at inference time. \textbf{Our augmented GPT-OSS-20B reaches 89\% execution accuracy, up from 76\% on the initial fine-tuning, and significantly up from its original 6.0\%.} \textit{In other words, our augmented fine-tuning with self-reflection enables GPT-OSS achieve an on par execution accuracy (89\%) compared to its frontier variant, GPT-5.2 (86\%)}. This is one of our main contributions.

However, solution accuracy remains flat. SLMs that fail to solve a problem on their first executable attempt do not succeed later. This suggests a separation between syntax errors (fixable through retry) and semantic errors (not fixable due to SLM capacity). Once the code compiles, the model has already committed to a constraint formulation. If that formulation is wrong, producing further executable code does not help.

\subsection{Ensemble}
Finally, as an alternative to self-reflection and instead of multiple attempts using \textit{the same SLM}, we evaluate a top-down ensemble (largest SLM first) to exploit the variation across SLMs. In our ensemble, models are tried in descending order of capability: GPT-OSS-20B $\rightarrow$ Gemma-2-9B $\rightarrow$ LLaMA-3.2-3B $\rightarrow$ LLaMA-3.2-1B $\rightarrow$ Qwen3-0.6B. If a model fails to produce executable code after retries, the next model is attempted. We consider a top-down strategy to increase the chance of getting an executable model where the solution accuracy remains highest (stronger models). Conversely, if we build this ensemble bottom-up (smallest SLM first) we risk obtaining an executing model when the solution reasoning capacity is lowest. 

Table~\ref{tab:ensemble} presents the main takeaways from our work \textsc{Learn2Zinc}. We start with the original GPT-OSS-20B, the largest of our SLMs, which has an extremely poor performance on generating \textsc{MiniZinc} models with only 6\% execution and 5\% solution accuracy. Next, our fine-tuning, the initial base strategy, \textsc{Learn2Zinc-Base}, detailed in \textsection \ref{sec:finetuning}, and its augmented version, \textsc{Learn2Zinc-Augmented} for error-correction detailed in \textsection \ref{sec:teaching} steadily improve \textsc{MiniZinc} coding abilities of GPT-OSS with up to 76\% execution accuracy. The solution accuracy also jumps from 5\% to 32\%. Our augmented version, \textsc{Learn2Zinc-Augmented}, has the additional property to serve well in self-reflection loops to fix its coding errors, which it is exactly trained for, that leads to another boost in execution accuracy to 89\%. \textbf{Finally, our ensemble of fined-tuned SLMS achieves a remarkable 98\% execution accuracy closing the syntactic aspect of the text-to-model translation task}. As a reference, the frontier GPT-5.2 achieves only 86\% execution accuracy for \textsc{MiniZinc} code with the best copilot, \textsc{Agentic + Code}, from \cite{text2model}. We show that it is possible to teach SLMs domain specific languages for text-to-model translation via fine-tuning. 

\textit{Overall, this is the central finding of our approach: with \textsc{Learn2Zinc-Augmented} fine-tuning, self-reflection, and SLM ensembles, the execution problem is effectively solved for \textsc{MiniZinc}. }Our method allows pushing  bottleneck to shift entirely to semantic reasoning about constraint modeling.

Beyond the syntax, what remains is the gap in solution accuracy for formal reasoning. Our SLM fine-tuning saturates around 35\% on solution accuracy. Even the frontier models such as GPT-5.2 struggles with this at 57\%. The best known solution on \textsc{IndustrOR} comes from ~\cite{lean-llm-opt} with 65\% using Gemini 3 Pro for generating \textsc{Gurobi} models in Python, as we discuss next. 


\begin{table}[t]
\centering
\begin{tabular}{llcc}
\toprule
\textbf{Model} & Approach & \textbf{Execution (\%)} & \textbf{Solution (\%)} \\
\midrule
GPT-OSS-20B & Original-4bit & 6.0 & 5.0 \\
GPT-OSS-20B & \textsc{Learn2Zinc-Base} & 66.0 & 27.0 \\
GPT-OSS-20B & \textsc{Learn2Zinc-Augmented} & 76.0 & 32.0 \\
GPT-OSS-20B & \textsc{Learn2Zinc-Aug+Self-Reflection} & 89.0 & 34.0 \\
Our fined-tuned SLMs & \textsc{Learn2Zinc-Ensemble} & \cellcolor{highlight}\best{98.0} & 35.0 \\
GPT-5.2 & \textsc{Agentic + Code}~\citep{text2model} & 86.0 & \cellcolor{highlight}\best{57.0} \\
\bottomrule
\end{tabular}
\caption{\textsc{Learn2Zinc} overall comparison with different strategies.}
\label{tab:ensemble}
\end{table}

\section{\textsc{Learn2Zinc} vs. Prior Work}
We compare \textsc{Learn2Zinc} with prior work on the same \textsc{IndustryOr} benchmark from  \textsc{Orlm}~\citep{orlm} and \textsc{Lean-Lllm-Opt}~\citep{lean-llm-opt}. Before numerical comparisons, let us first highlight that such a comparison is subject to several caveats and should be treated \textit{only directionally}.

First, target constraint languages differ. \textsc{Orlm}~\citep{orlm} generates Python code for the \textsc{Copt} solver whereas \textsc{Lean-Lllm-Opt}~\citep{lean-llm-opt} generates Python code for the \textsc{Gurobi} solver. 

Python is a general-purpose programming language and is abundant in the LLM pretraining data. As such, LLMs produce syntactically valid code easily for Python. Contrarily, our approach generates \textsc{MiniZinc}, a domain-specific modeling language, and as shown in our experiments, remains out-of-distribution, especially for SLMs. An important contribution of \textsc{Learn2Zinc} is to address syntax challenge which is taken for granted in Pythonic approaches.

Second, evaluation protocols differ. \textsc{Orlm} reports Pass@k by sampling k independent outputs and counting success if \textit{any} one is correct. Our evaluation protocol reflects an agentic loop: we iterate until execution succeeds or a budget is exhausted, while sharing error messages in between, and counting the total number of LLM calls in average termination. These results are not directly comparable, but we report both for context.

Third, the underlying LLMs differ. \textsc{Orlm} fine-tunes over LLaMA-3-8B, similar but not exactly identical to our case, while \textsc{Lean-Lllm-Opt} is based on GPT-4.1 and GPT-OSS-20B. For a complete picture, we also include results on GPT and Gemini for generating Gurobi-Python from ~\citep{lean-llm-opt}, as-is, as well as results on GPT for generating \textsc{MiniZinc} from~\citep{text2model} as-is.

Table~\ref{tab:IndustryOr_comparison} shows results on \textsc{IndustryOr} benchmark for \textsc{Orlm}~\citep{orlm}, \textsc{Lean-Lllm-Opt}~\citep{lean-llm-opt}, \textsc{Text2Model}~\citep{text2model}, and \textsc{Learn2Zinc}. Our best fine-tuned GPT-OSS-20B achieves 34\% on \textsc{IndustryOr}, below the Pythonic methods. This gap reflects two factors. First, \textsc{MiniZinc} syntax remains harder than Python even though \textsc{Learn2Zinc-Augmented} can address it. Second, our SLM are lighter than the frontier models used by other methods. The \textsc{Text2Model} result (57\% with GPT-5.2) shows that stronger models with \textsc{MiniZinc} can approach Pythonic performance.

\begin{table}[t]
\centering
\begin{tabular}{@{}llcc@{}}
\toprule
\textbf{Method} & \textbf{Model} & \textbf{LLM Calls} & \textbf{\textsc{IndustryOr} (\%)} \\
\midrule
\multicolumn{4}{l}{\textit{Fine-tuning approach for \textsc{Copt} in Python}} \\
ORLM (Pass@1) & LLaMA-3-8B & 1 & 38.0 \\
ORLM (Pass@8) & LLaMA-3-8B & 8 &  \cellcolor{highlight}\textbf{49.0} \\
\midrule
\multicolumn{4}{l}{\textit{Agentic framework for \textsc{Gurobi} in Python}} \\
\textsc{Lean-Lllm-Opt}$^\dagger$ & GPT-OSS-20B & 5+ & 59.0 \\
\textsc{Lean-Lllm-Opt}$^\dagger$ & GPT-4.1 & 5+ & \cellcolor{highlight}\textbf{65.0} \\
\midrule
\multicolumn{4}{l}{\textit{Direct prompting for \textsc{Gurobi} in Python}$^\dagger$} \\
-- & GPT-OSS-20B & 1& 42.0 \\
-- & GPT-4.1 & 1& 54.0 \\
-- & GPT-5 & 1& 58.0 \\
-- & GPT-5.2 & 1 & 56.0 \\
-- & Gemini 3 Pro & 1&  \cellcolor{highlight}\textbf{65.0} \\
\midrule
\multicolumn{4}{l}{\textit{\textsc{Text2Model} copilots for \textsc{MiniZinc}}} \\
\textsc{Text2Model: \textsc{Agentic + Code}} & GPT-OSS-20B & 1 & 5.0 \\
\textsc{Text2Model: \textsc{Agentic + Code}} & GPT-5.2 & 1 & 49.0 \\
\textsc{Text2Model: \textsc{Agentic + Code}} & GPT-5.2 & 5 & \cellcolor{highlight}\textbf{57.0} \\
\midrule
\multicolumn{4}{l}{\textit{Fine-tuning \textsc{MiniZinc} (this paper)}} \\
\textsc{Learn2Zinc-Augmented} & GPT-OSS-20B & 1 & 32.0 \\
\textsc{Learn2Zinc-Augmented+Self-Reflection} & GPT-OSS-20B & 1.62 & \cellcolor{highlight}\textbf{34.0} \\
\bottomrule
\end{tabular}
\caption{\textsc{IndustryOr} comparisons across \textsc{Orlm}, generating \textsc{Copt} models in Python, \textsc{Lean-Lllm-Opt} generating \textsc{Gurobi} models in Python, \textsc{Text2Model} and \textsc{Learn2Zinc} generating \textsc{MiniZinc} across different LLMs and SLMs. Results marked with $\dagger$ are from \citet{lean-llm-opt}.}
\label{tab:IndustryOr_comparison}
\end{table}

The value of \textsc{Learn2Zinc} is complementary and methodological: we produce solver- and paradigm- agnostic \textsc{MiniZinc} models that can solve optimization \textit{and} satisfaction problems, whereas \textsc{Orlm} is tied to \textsc{Copt} and \textsc{Lean-Lllm-Opt} is tied to \textsc{Gurobi}. We maintain solver and paradigm flexibility while addressing SLMs's disadvantage on domain-specific languages via \textsc{Learn2Zinc} fine-tuning.


\vspace{-0.2cm}
\section{Discussions \& Error Analysis}
\label{sec:discussions}
\vspace{-0.2cm}

Our immediate obsersation is that \textbf{out-of-the-box SLMs achieve near 0\% performance}. Three factors combine to make \textsc{MiniZinc} out-of-distribution for current LLMs. First, \textsc{MiniZinc} has minimal GitHub presence compared to mainstream languages. Second, it uses a declarative paradigm unlike the imperative code that makes up most pretraining data. Third, no syntactically similar language exists for transfer learning. The combination of rare syntax and unfamiliar paradigm explains the complete failure of unfinetuned models.

We also notice that \textbf{CoT is not effective when fine-tuning SLMs} as \textsc{Learn2Zinc-CoT} underperforms. When prompting an LLM, CoT often helps because models already know the target language and they benefit from structured reasoning but fine-tuning is different. \textsc{Learn2Zinc-CoT} requires the SLM to reason about valid \textsc{MiniZinc} constructs it has seldom or never seen. SLMs might not reason about syntax that they have not acquired. The reasoning traces may actually confuse the model and introduces error. That said, the errors we observe suggest that a different kind of \textsc{Learn2Zinc-CoT} could help. Specifically, \textsc{Learn2Zinc-CoT} focused on mathematical and constraint-level reasoning rather than simple step-by-step narration may reduce the semantic failures that current models make. This is one possible future study. 

Regarding execution accuracy, \textbf{error-correction is clearly helpful for fine-tuning}. The \textsc{Learn2Zinc-Augmented} strategy directly targets the main bottleneck of syntax errors. By training on realistic error-correction examples collected through cross-model bootstrapping, S:< learn to \textit{both generate code and fix common mistakes}. Using errors from multiple model sizes gives broad coverage of failure patterns. It would be interesting to curate such fine-tuning datasets for domain-specific languages. 

Regarding solution accuracy, our ensemble that has 98\% execution accuracy only achieves 35\% solution accuracy. This gap shows a clear split: \textbf{while syntax is learnable, but constraint modeling remains hard for SLMs}. The models produce valid \textsc{MiniZinc} code that compiles and runs, but often get the problem semantics wrong. In our error anaylsis  find two main categories of semantic error:

\textbf{Type I Error: Contradictory constraints causing infeasibility.}
Models introduce auxiliary variables with two incompatible definitions. For example, a variable may be constrained to equal both the leftover supply and the truck count computed from the same shipped quantity. The solver returns unsat even though the underlying problem is perfectly feasible. An example of this type of error is given in Appendix~\ref{app:unsat-example}. As shown in Appendix-Table~\ref{tab:constraint-errors} (Type~I), fixing these contradictions alone could improve solution accuracy by up to 18 percentage points depending on the model. 

\textbf{Type II Error: Phantom variables forcing zero objectives.}
Models create unnecessary decision variables (for example, binary indicators in a pure LP) and constrain them to equal expressions that exceed their declared bounds. The only way to satisfy these constraints is to set all decision variables to zero, which gives a trivially wrong objective.  An example of this type of error is given in Appendix~\ref{app:phantom-example}. Appendix-Table~\ref{tab:constraint-errors} (Type~II) shows this affects up to 20 percentage points. 

\section{Related Work}
\label{sec:related}

There is growing interest in leveraging LLMs for optimization tasks to transform the way that decision makers interact with solvers in the form of a co-pilot system~\citep{SimchiLevi2025Democratizing,Wasserkrug2025,text2model,tsouros2023holy}.

\paragraph{From Natural Language to Optimization Models:}
Earlier systems relied on rule-based parsing to construct mixed-integer or logic models (e.g., \textsc{LGPSolver} for logic-grid puzzles~\citep{jabrayilzade2020lgpsolver} and \textsc{AutoLP} for linear programs~\citep{islam2021autoLP}). The \textsc{\textsc{Nl4Opt}} competition~\citep{lpwp} formalized the task as a two-step pipeline: named-entity recognition followed by code generation. Follow-up works such as \textsc{LaTeX2Solver}~\citep{ramamonjison2023latex2solver} extended parsing to mathematical documents, while the ``Holy Grail 2.0'' blueprint~\citep{tsouros2023holy} envisioned conversational assistants that refine models interactively.

\paragraph{Prompt-Driven and Learning-based Copilots:}
The \textsc{Ner4Opt} line of works~\citep{ner4opt2023,ner4opt2024} showed that fine-tuning transformers on optimization-specific corpora and inline entity tags boosts accuracy. Retrieval-augmented prompting specific to Constraint Programming settings has also proven effective~\citep{michailidis_et_al:LIPIcs.CP.2024.20}. Modular agent pipelines push this further: \textsc{ComplexOR} employs a chain-of-experts architecture for difficult OR problems~\citep{complexor}, while \textsc{OptiMUS} decomposes formulation, debugging, and solving into separate GPT agents~\citep{optimus}. The recent \textsc{Gala} framework builds global agents for CP~\citep{cai2025galagloballlmagents}. Fine-tuning approaches include \textsc{Orlm}~\citep{orlm} and \textsc{LllmOpt}~\citep{jiang2025llmopt}.

\paragraph{Code Generation for Rare Languages:}
EsoLang-Bench~\citep{esolangbench} shows frontier models drop from 85-95\% to 0-11\% on rare languages, with few-shot and self-reflection providing negligible benefit when pretraining coverage is absent. Our work differs in that we study whether fine-tuning, rather than prompting, can close this gap for constraint programming.

\paragraph{Datasets and Evaluation:}
Optimization-centric corpora include \textsc{\textsc{Nl4Opt}}~\citep{lpwp}, \textsc{Nlp4Lp}~\citep{optimus}, \textsc{\textsc{IndustryOr}}~\citep{orlm}, and \textsc{MILP} synthesis datasets~\citep{li2023synthesizingmixedintegerlinearprogramming}. Recent datasets include \textsc{Planetarium}~\citep{zuo2025planetariumrigorousbenchmarktranslating} for planning domain definition language (PDDL), \textsc{Ehop}~\citep{duchnowski2025ehopdataseteverydaynphard} for everyday NP-Hard problems, and \textsc{DualSchool}~\citep{klamkin2025dualschoolreliablellmsoptimization} for optimization education. Evaluation metrics have evolved from exact string match to execution and solution accuracy.

Our work on \textsc{Learn2Zinc} builds on the \textsc{Text2Zinc} dataset~\cite{text2zinc} and \textsc{Text2Model} copilots\cite{text2model}, which together provide the data and baselines for the fine-tuning study presented here. The closest to our approach on \textsc{Learn2Zinc} is the work on \textsc{OptiMind}\cite{zhang2026optimindteachingllmsthink} which fine-tunes GPT-OSS-20B on optimization modeling for Gurobi modeling in Python. Interestingly, \cite{text2model} reports that the \textsc{MiniZinc} capabilities of \textsc{OptiMind} fine-tuned GPT-OSS-20B is reduced to zero. This is a call for future research on fine-tuning models with potentially unintended consequences on losing capabilities in other domains. 

\section{Conclusion}
\label{sec:conclusion}

We presented \textsc{Learn2Zinc}, \textbf{the first systematic study of fine-tuning small language models (SLMs) for generating \textsc{MiniZinc} constraint models} from natural language. Our results show that \textsc{MiniZinc} is genuinely out-of-distribution for current models, with near-zero execution accuracy without fine-tuning. This highlights the limitations of prompt-based approaches for domain-specific languages with limited pretraining coverage.

To address this, we introduced \textbf{several fine-tuning strategies} including cross-model error bootstrapping, a method for constructing realistic syntax error-correction datasets by leveraging failures across multiple models. Combined with grammar-based synthetic corruptions, this approach enables models to learn both code generation and targeted error correction. We further show that, unlike in prompting settings, chain-of-thought fine-tuning does not improve performance when foundational syntax knowledge is absent.

Our \textsc{Learn2Zinc-Augmented} strategy significantly improves execution accuracy across all model sizes of SLMs. When combined with self-reflection and a simple top-down ensemble, \textbf{we achieve 98\% execution accuracy, effectively resolving the syntax bottleneck for \textsc{MiniZinc}}. However, solution accuracy saturates at 34\%, revealing a clear gap between syntactic correctness and semantic reasoning. Our initial error analysis attributes this gap to recurring issues such as contradictory constraints and phantom variables. These findings suggest that while syntax can be learned efficiently through targeted fine-tuning, constraint reasoning remains the primary challenge for SLM-based text-to-model translation.

Future work should focus on \textbf{closing the execution-solution gap} through improved supervision and reasoning support. Promising directions include training with structured reasoning traces aligned with constraint modeling subtasks, leveraging intermediate supervision for \textsc{MiniZinc} generation, and distilling reasoning capabilities from larger models. In addition, integrating fine-tuned models with grammar-constrained decoding, prompting strategies, and modular or agentic pipelines may further improve solution quality.

\bibliography{references}

@InProceedings{minizinc,
author="Nethercote, Nicholas
and Stuckey, Peter J.
and Becket, Ralph
and Brand, Sebastian
and Duck, Gregory J.
and Tack, Guido",
editor="Bessi{\`e}re, Christian",
title="MiniZinc: Towards a Standard CP Modelling Language",
booktitle="Principles and Practice of Constraint Programming -- CP 2007",
year="2007",
publisher="Springer Berlin Heidelberg",
address="Berlin, Heidelberg",
pages="529--543",
isbn="978-3-540-74970-7"
}

@inproceedings{guns2019increasing,
    title={Increasing modeling language convenience with a universal n-dimensional array, CPpy as python-embedded example},
    author={Guns, Tias},
    booktitle={Proceedings of the 18th workshop on Constraint Modelling and Reformulation at CP (Modref 2019)},
    volume={19},
    year={2019}
}

@inbook{Bussieck2004,
	address = {Boston, MA},
	author = {Bussieck, Michael R. and Meeraus, Alex},
	booktitle = {Modeling Languages in Mathematical Optimization},
	doi = {10.1007/978-1-4613-0215-5_8},
	isbn = {978-1-4613-0215-5},
	pages = {137--157},
	publisher = {Springer US},
	title = {General Algebraic Modeling System (GAMS)},
	year = {2004}
}

@misc{openai2024gpt4technicalreport,
      title={GPT-4 Technical Report}, 
      author={OpenAI and others},
      year={2024},
      eprint={2303.08774},
      archivePrefix={arXiv},
      primaryClass={cs.CL},
      url={https://arxiv.org/abs/2303.08774}
}

@misc{geminiteam2025geminifamilyhighlycapable,
      title={Gemini: A Family of Highly Capable Multimodal Models}, 
      author={Gemini Team and others},
      year={2025},
      eprint={2312.11805},
      archivePrefix={arXiv},
      primaryClass={cs.CL},
      url={https://arxiv.org/abs/2312.11805}
}

@misc{deepseekai2025deepseekv3technicalreport,
      title={DeepSeek-V3 Technical Report}, 
      author={DeepSeek-AI and others},
      year={2025},
      eprint={2412.19437},
      archivePrefix={arXiv},
      primaryClass={cs.CL},
      url={https://arxiv.org/abs/2412.19437}
}

@misc{lpwp,
  doi = {10.48550/ARXIV.2209.15565},
  url = {https://arxiv.org/abs/2209.15565},
  author = {Ramamonjison, Rindranirina and Li, Haley and Yu, Timothy T. and He, Shiqi and Rengan, Vishnu and Banitalebi-Dehkordi, Amin and Zhou, Zirui and Zhang, Yong},
  title = {Augmenting Operations Research with Auto-Formulation of Optimization Models from Problem Descriptions},
  publisher = {arXiv},
  year = {2022}
}

@inproceedings{complexor, 
  title={Chain-of-Experts: When LLMs Meet Complex Operations Research Problems},
  author={Xiao, Ziyang and Zhang, Dongxiang and Wu, Yangjun and Xu, Lilin and Wang, Yuan Jessica and Han, Xiongwei and Fu, Xiaojin and Zhong, Tao and Zeng, Jia and Song, Mingli and others},
  booktitle={The Twelfth International Conference on Learning Representations},
  year={2023}
}

@misc{
csplib,
title={{CSPLib}: A problem library for constraints},
author={Jefferson, Christopher and Miguel, Ian and Hnich, Brahim and Walsh, Toby and Gent, Ian P.},
year= {1999},
url={https://openreview.net/forum?id=HobyL1B9CZ}
}

@misc{optimus,
      title={OptiMUS-0.3: Using Large Language Models to Model and Solve Optimization Problems at Scale}, 
      author={Ali AhmadiTeshnizi and Wenzhi Gao and Herman Brunborg and Shayan Talaei and Madeleine Udell},
      year={2024},
      eprint={2407.19633},
      archivePrefix={arXiv},
      primaryClass={cs.AI},
      url={https://arxiv.org/abs/2407.19633}
}

@misc{hakank,
author       = {Kjellerstrand, H{\aa}kan},
  title        = {Hakank's Page},
  year         = {2025},
  note         = {GitHub repository, Retrieved February 13th, 2025},
  howpublished = {\url{https://github.com/hakank/hakank}}
}

@misc{mamo,
      title={Mamo: a Mathematical Modeling Benchmark with Solvers}, 
      author={Xuhan Huang and Qingning Shen and Yan Hu and Anningzhe Gao and Benyou Wang},
      year={2024},
      eprint={2405.13144},
      archivePrefix={arXiv},
      primaryClass={cs.AI},
      url={https://arxiv.org/abs/2405.13144}
}

@misc{orlm,
      title={ORLM: A Customizable Framework in Training Large Models for Automated Optimization Modeling}, 
      author={Chenyu Huang and Zhengyang Tang and Dongdong Ge and Shixi Hu and Ruoqing Jiang and Benyou Wang and Zizhuo Wang and Xin Zheng},
      year={2024},
      eprint={2405.17743},
      archivePrefix={arXiv},
      primaryClass={cs.CL},
      url={https://arxiv.org/abs/2405.17743}
}

@article{ner4opt2024,
  title={Ner4Opt: named entity recognition for optimization modelling from natural language},
  author={Kad{\i}o{\u{g}}lu, Serdar and Pravin Dakle, Parag and Uppuluri, Karthik and Politi, Regina and Raghavan, Preethi and Rallabandi, SaiKrishna and Srinivasamurthy, Ravisutha},
  journal={Constraints},
  pages={1--39},
  year={2024},
  publisher={Springer}
}

@inproceedings{ner4opt2023,
  title={Ner4opt: Named entity recognition for optimization modelling from natural language},
  author={Dakle, Parag Pravin and Kad{\i}o{\u{g}}lu, Serdar and Uppuluri, Karthik and Politi, Regina and Raghavan, Preethi and Rallabandi, SaiKrishna and Srinivasamurthy, Ravisutha},
  booktitle={International Conference on Integration of Constraint Programming, Artificial Intelligence, and Operations Research},
  pages={299--319},
  year={2023},
  organization={Springer}
}

@article{SimchiLevi2025Democratizing,
  author    = {David Simchi-Levi and Tinglong Dai and Ishai Menache and Michelle Xiao Wu},
  title     = {Democratizing Optimization with Generative AI},
  journal   = {Johns Hopkins Carey Business School Research Paper Forthcoming},
  year      = {2025},
  month     = {October 24}
}

@article{Wasserkrug2025,
  author       = {Segev Wasserkrug and Leonard Boussioux and Dick den Hertog and Farzaneh Mirzazadeh and S. Ilker Birbil and Jannis Kurtz and Donato Maragno},
  title        = {Enhancing Decision Making Through the Integration of Large Language Models and Operations Research Optimization},
  journal      = {Proceedings of the AAAI Conference on Artificial Intelligence},
  year         = {2025},
  volume       = {39},
  number       = {27},
  pages        = {28643--28650},
  doi          = {10.1609/aaai.v39i27.35090}
}

@article{tsouros2023holy,
  title={Holy Grail 2.0: From Natural Language to Constraint Models},
  author={Tsouros, Dimos and Verhaeghe, Helene and Kad{\i}o{\u{g}}lu, Serdar and Guns, Tias},
  journal={arXiv preprint arXiv:2308.01589},
  year={2023}
}

@misc{cai2025galagloballlmagents,
      title={Gala: Global LLM Agents for Text-to-Model Translation}, 
      author={Junyang Cai and Serdar Kad{\i}o{\u{g}}lu and Bistra Dilkina},
      year={2025},
      eprint={2509.08970},
      archivePrefix={arXiv},
      primaryClass={cs.AI},
      url={https://arxiv.org/abs/2509.08970}
}

@inproceedings{jiang2025llmopt,
  title     = {LLMOPT: Learning to Define and Solve General Optimization Problems from Scratch},
  author    = {Caigao Jiang and Xiang Shu and Hong Qian and Xingyu Lu and Jun Zhou and Aimin Zhou and Yang Yu},
  booktitle = {Proceedings of the 13th International Conference on Learning Representations (ICLR)},
  year      = {2025},
  address   = {Singapore},
  url       = {https://openreview.net/forum?id=9OMvtboTJg}
}

@InProceedings{michailidis_et_al:LIPIcs.CP.2024.20,
  author =	{Michailidis, Kostis and Tsouros, Dimos and Guns, Tias},
  title =	{{Constraint Modelling with LLMs Using In-Context Learning}},
  booktitle =	{30th International Conference on Principles and Practice of Constraint Programming (CP 2024)},
  pages =	{20:1--20:27},
  series =	{Leibniz International Proceedings in Informatics (LIPIcs)},
  year =	{2024},
  volume =	{307},
  publisher =	{Schloss Dagstuhl -- Leibniz-Zentrum f{\"u}r Informatik},
  doi =		{10.4230/LIPIcs.CP.2024.20}
}

@inproceedings{jabrayilzade2020lgpsolver,
  title     = {{LGPS}olver -- Solving Logic Grid Puzzles Automatically},
  author    = {Jabrayilzade, Elgun and Tekir, Selma},
  booktitle = {Findings of the Association for Computational Linguistics: EMNLP 2020},
  pages     = {1118--1123},
  year      = {2020},
  month     = nov,
  publisher = {Association for Computational Linguistics},
  doi       = {10.18653/v1/2020.findings-emnlp.100}
}

@inproceedings{islam2021autoLP,
  title     = {Automatic Formulation and Optimization of Linear Problems from a Structured Paragraph},
  author    = {Islam, Md. Shahidul and Mamud, Fariha and Haque, Rakib Ul and Saber, Ahmed Y.},
  booktitle = {Proc.\ 2021 Int. Conf. on Science \& Contemporary Technologies (ICSCT)},
  year      = {2021},
  doi       = {10.1109/ICSCT53883.2021.9642516}
}

@inproceedings{ramamonjison2023latex2solver,
    title = "{L}a{T}e{X}2{S}olver: a Hierarchical Semantic Parsing of {L}a{T}e{X} Document into Code for an Assistive Optimization Modeling Application",
    author = "Ramamonjison, Rindra and Yu, Timothy and Xing, Linzi and Mostajabdaveh, Mahdi and Li, Xiaorui and Fu, Xiaojin and Han, Xiongwei and Chen, Yuanzhe and Li, Ren and Mao, Kun and Zhang, Yong",
    booktitle = "Proceedings of the 61st Annual Meeting of the Association for Computational Linguistics (Volume 3: System Demonstrations)",
    month = jul,
    year = "2023",
    publisher = "Association for Computational Linguistics",
    doi = "10.18653/v1/2023.acl-demo.45",
    pages = "471--478"
}

@misc{li2023synthesizingmixedintegerlinearprogramming,
      title={Synthesizing mixed-integer linear programming models from natural language descriptions}, 
      author={Qingyang Li and Lele Zhang and Vicky Mak-Hau},
      year={2023},
      eprint={2311.15271},
      archivePrefix={arXiv},
      primaryClass={math.OC},
      url={https://arxiv.org/abs/2311.15271}
}

@misc{zuo2025planetariumrigorousbenchmarktranslating,
      title={Planetarium: A Rigorous Benchmark for Translating Text to Structured Planning Languages}, 
      author={Max Zuo and Francisco Piedrahita Velez and Xiaochen Li and Michael L. Littman and Stephen H. Bach},
      year={2025},
      eprint={2407.03321},
      archivePrefix={arXiv},
      primaryClass={cs.CL},
      url={https://arxiv.org/abs/2407.03321}
}

@misc{duchnowski2025ehopdataseteverydaynphard,
      title={EHOP: A Dataset of Everyday NP-Hard Optimization Problems}, 
      author={Alex Duchnowski and Ellie Pavlick and Alexander Koller},
      year={2025},
      eprint={2502.13776},
      archivePrefix={arXiv},
      primaryClass={cs.CL},
      url={https://arxiv.org/abs/2502.13776}
}

@misc{klamkin2025dualschoolreliablellmsoptimization,
      title={DualSchool: How Reliable are LLMs for Optimization Education?}, 
      author={Michael Klamkin and Arnaud Deza and Sikai Cheng and Haoruo Zhao and Pascal Van Hentenryck},
      year={2025},
      eprint={2505.21775},
      archivePrefix={arXiv},
      primaryClass={cs.LG},
      url={https://arxiv.org/abs/2505.21775}
}

@misc{esolangbench,
      title={EsoLang-Bench: Evaluating Genuine Reasoning in Large Language Models via Esoteric Programming Languages}, 
      author={Aman Sharma and Paras Chopra},
      year={2026},
      eprint={2603.09678},
      archivePrefix={arXiv},
      primaryClass={cs.AI},
      url={https://arxiv.org/abs/2603.09678}, 
}

@misc{lean-llm-opt,
      title={Large-Scale Optimization Model Auto-Formulation: Harnessing LLM Flexibility via Structured Workflow}, 
      author={Kuo Liang and Yuhang Lu and Jianming Mao and Shuyi Sun and Chunwei Yang and Congcong Zeng and Xiao Jin and Hanzhang Qin and Ruihao Zhu and Chung-Piaw Teo},
      year={2026},
      eprint={2601.09635},
      archivePrefix={arXiv},
      primaryClass={cs.AI},
      url={https://arxiv.org/abs/2601.09635}, 
}

@misc{text2zinc,
      title={Text2Zinc: A Cross-Domain Dataset for Modeling Optimization and Satisfaction Problems in MiniZinc}, 
      author={Akash Singirikonda and Serdar Kad{\i}o{\u{g}}lu and Karthik Uppuluri},
      year={2025},
      eprint={2503.10642},
      archivePrefix={arXiv},
      primaryClass={cs.CL},
      url={https://arxiv.org/abs/2503.10642}, 
}

@misc{text2model,
      title={Modeling Copilots for Text-to-Model Translation}, 
      author={Serdar Kad{\i}o{\u{g}}lu and Karthik Uppuluri and Akash Singirikonda},
      year={2026},
      eprint={2604.12955},
      archivePrefix={arXiv},
      primaryClass={cs.AI},
      url={https://arxiv.org/abs/2604.12955}, 
}

@misc{zhang2026optimindteachingllmsthink,
      title={OptiMind: Teaching LLMs to Think Like Optimization Experts}, 
      author={Xinzhi Zhang and Zeyi Chen and Humishka Zope and Hugo Barbalho and Konstantina Mellou and Marco Molinaro and Janardhan Kulkarni and Ishai Menache and Sirui Li},
      year={2026},
      eprint={2509.22979},
      archivePrefix={arXiv},
      primaryClass={cs.LG},
      url={https://arxiv.org/abs/2509.22979}, 
}
\bibliographystyle{iclr2026_conference}

\newpage

\appendix
\section{Fine-tuning Examples}
\label{appendix:examples}
\subsection*{\ptc{} Example}

\textit{Instruction:}\newline\newline
Generate \textsc{MiniZinc} code for the following optimization problem.\newline\newline
A rescue material distribution company needs to allocate rescue materials to three disaster-stricken areas: A, B, and C. These places have different degrees of disaster, so the required amount of rescue materials is also different. The goal is to minimize the total distribution time associated with these locations, with a distribution time weight of 50 hours for A, 100 hours for B, and 200 hours for C.\newline\newline
The allocation must adhere to the following conditions:\newline\newline
- The total amount of rescue materials for locations A and B cannot exceed 500 units.\newline
- The total amount of rescue materials for locations B and C must be at least 300 units.\newline
- The amount of rescue materials for location C cannot exceed the maximum demand of 600 units.\newline
- The amount of rescue materials for location A must be at least 50 units.\newline\newline
\textit{Response:}
\begin{lstlisting}[style=minizinc]
% Parameters (embed all data values directly here)
int: wA = 50;
int: wB = 100;
int: wC = 200;

int: A_min = 50;
int: A_max = 500;

int: B_min = 0;
int: B_max = 400;

int: C_min = 0;
int: C_max = 600;

int: AB_max = 500;
int: BC_min = 300;

% Variables
var A_min..A_max: A;
var B_min..B_max: B;
var C_min..C_max: C;

var 0..(wA*A_max + wB*B_max + wC*C_max): total_time;

% Constraints
constraint A + B <= AB_max;
constraint B + C >= BC_min;

constraint total_time = wA*A + wB*B + wC*C;

% Objective
solve minimize total_time;
\end{lstlisting}

\subsection*{\ptcr{} Example}

\textit{Instruction:}\newline\newline
Generate MiniZinc code for the following optimization problem. \textbf{First outline your reasoning, then provide the code.}\newline\newline
A rescue material distribution company needs to allocate rescue materials to three disaster-stricken areas: A, B, and C. These places have different degrees of disaster, so the required amount of rescue materials is also different. The goal is to minimize the total distribution time associated with these locations, with a distribution time weight of 50 hours for A, 100 hours for B, and 200 hours for C.\newline\newline
The allocation must adhere to the following conditions:\newline\newline
- The total amount of rescue materials for locations A and B cannot exceed 500 units.\newline
- The total amount of rescue materials for locations B and C must be at least 300 units.\newline
- The amount of rescue materials for location C cannot exceed the maximum demand of 600 units.\newline
- The amount of rescue materials for location A must be at least 50 units.\newline\newline
Considering that each location has a specific demand for rescue materials (A: 0-500, B: 0-400, C: 0-600), and due to the limitation of the quantity of materials, since we cannot have fractional units of rescue materials, what is the minimum total distribution time in hours for the optimal allocation of rescue materials to the three locations within the specified demand ranges? Please round the answer to the nearest hour.\newline\newline

\textit{Response:}\newline\newline

\textbf{Reasoning:}\newline\newline
Variables: A $\in$ [50, 500], B $\in$ [0, 400], C $\in$ [0, 600]\newline\newline 
Constraints: A + B $\leq$ 500, B + C $\geq$ 300\newline\newline
Objective: minimize total\_time = 50*A + 100*B + 200*C\newline\newline

\textbf{Code:}
\begin{lstlisting}[style=minizinc]
% Parameters (embed all data values directly here)
int: wA = 50;
int: wB = 100;
int: wC = 200;

int: A_min = 50;
int: A_max = 500;

int: B_min = 0;
int: B_max = 400;

int: C_min = 0;
int: C_max = 600;

int: AB_max = 500;
int: BC_min = 300;

% Variables
var A_min..A_max: A;
var B_min..B_max: B;
var C_min..C_max: C;

var 0..(wA*A_max + wB*B_max + wC*C_max): total_time;

% Constraints
constraint A + B <= AB_max;
constraint B + C >= BC_min;

constraint total_time = wA*A + wB*B + wC*C;

% Objective
solve minimize total_time;
\end{lstlisting}

\subsection*{Syntax Correction Example}

\textit{Instruction:}\newline\newline
Fix the following MiniZinc code that has syntax errors.\newline\newline
Problem: A shipping company needs to transport goods from a source port to a destination port. There are 3 ships available for selection: Ship A, B, and C. The cargo capacity, transportation cost, and transportation time for each ship are shown in the table below:\newline

\begin{table}[H]
\centering
\begin{tabular}{lccc}
\toprule
Ship & Capacity (tons) & Cost (M yuan) & Time (days) \\
\midrule
A & 1000 & 10 & 10 \\
B & 800  & 8  & 8  \\
C & 1200 & 12 & 7  \\
\bottomrule
\end{tabular}
\end{table}

Now, there is a need to transport 3000 tons of cargo and it must arrive within 15 days. How should the ships be scheduled for transportation to minimize the total transportation cost? Each ship can only make one trip, and the amount of cargo transported cannot exceed its cargo capacity.\newline\newline

\textit{Broken code:}
\begin{lstlisting}[style=minizinc]
% Parameters (embed all data values directly here)
int: NSHIPS = 3;
set of int: SHIPS = 1..NSHIPS;

array[SHIPS] of string: ship_name = ["A","B","C"];

% Cargo capacities in tons
array[SHIPS] of int: cap = [1000, 800, 1200];

% Transportation costs (million yuan)
array[SHIPS] of int: cost = [10, 8, 12];

% Transportation times (days)
array[SHIPS] of int: time = [10, 8, 7];

int: demand = 3000;                 % tons
int: max_days = 15;

% Upper bounds for number of trips per ship
int: max_trips = ceil(demand div min(cap));   % 3 trips needed

% Variables
array[SHIPS] of var 0..max_trips: trips;

var 0..demand: cargo_each;                 % tons per ship
array[SHIPS] of var 0..demand: x;           % tons shipped per ship
var 0..(max_trips * max(cost)): total_cost;

% Constraints
constraint forall(i in SHIPS) (
  cargo_each[i] <= cap[i] /\
  x[i] = trips[i] * cargo_each[i] /\
  x[i] <= cap[i] /\
  x[i] >= 0
);

constraint sum(i in SHIPS)(x[i]) = demand;

% Ship capacity limits (redundant but explicit)
constraint forall(i in SHIPS) (
  cargo_each[i] <= cap[i] /\
  x[i] <= cap[i] /\
  x[i] >= 0
);

% Time window: all ships must be used (each ship makes one trip)
constraint forall(i in SHIPS) (
  time[i] <= max_days /\
  time[i] >= 0
);

constraint total_cost = sum(i in SHIPS)(cost[i] * trips[i]);

% Objective
solve minimize total_cost;
\end{lstlisting}

\textit{Error:} \texttt{type error: array access attempted on expression of type `var int'}
\newline\newline

\textit{Response:}
\begin{lstlisting}[style=minizinc]
% Parameters (embed all data values directly here)
int: N = 3;
set of int: SHIPS = 1..N;

array[SHIPS] of string: name = ["A","B","C"];

array[SHIPS] of int: capacity = [1000, 800, 1200];          % tons
array[SHIPS] of int: cost = [10, 8, 12];                    % million yuan
array[SHIPS] of int: time = [10, 8, 7];                     % days

int: demand = 3000; % tons
int: max_days = 15;

% Variables
array[SHIPS] of var 0..1: make_trip;
array[SHIPS] of var 0..demand: cargo;

var 0..(max_days * max(cost)): total_cost;

% Constraints
constraint forall(i in SHIPS) (
  cargo[i] <= capacity[i] * make_trip[i]
);

constraint sum(i in SHIPS)(cargo[i]) = demand;

constraint forall(i in SHIPS) (
  make_trip[i] = bool2int(time[i] >= 1)
);

constraint sum(i in SHIPS)(time[i] * make_trip[i]) <= max_days;

constraint total_cost = sum(i in SHIPS)(cost[i] * make_trip[i]);

% Objective
solve minimize total_cost;
\end{lstlisting}

\newpage

\section{Training Details}
\label{app:training_details}
 
 
 
\begin{table}[h!]
\centering
\caption{Training hyperparameters for each model. All models used max sequence length of 4,096 tokens and batch size of 2.}
\label{tab:training_config}
\begin{tabular}{lrrrrr}
\toprule
Model & LoRA $r$ & LoRA $\alpha$ & LR & Epochs & Quantization \\
\midrule
Qwen3-0.6B & 64 & 64 & 2e-4 & 3 & 8-bit \\
LLaMA-3.2-1B & 64 & 64 & 2e-4 & 3 & 8-bit \\
LLaMA-3.2-3B & 64 & 64 & 2e-4 & 3 & 8-bit \\
Gemma-2-9B & 64 & 64 & 1e-4 & 3 & 8-bit \\
GPT-OSS-20B & 32 & 64 & 2e-4 & 3 & 4-bit \\
\bottomrule
\end{tabular}
\end{table}

\clearpage
\section{Error Analysis}
\label{app:error_analysis}
\subsection{Type I: Contradictory Constraints}
\label{app:unsat-example}
\textbf{\large LLM-Generated Constraint Errors in MiniZinc}
\vspace{8pt}

{\small\textbf{Problem Description.} The full problem statement (Italian container transportation, 6 warehouses to 5 ports) can be found at: \url{https://huggingface.co/datasets/CardinalOperations/IndustryOr/viewer/default/test?q=transportation+cost&row=6}}

\vspace{2pt}

\begin{minipage}[t]{0.485\textwidth}
\begin{tcolorbox}[
  title={\small\bfseries\sffamily (a) LLM-Generated Code \\ Result: UNSATISFIABLE},
  colback=bugbg, colframe=bugframe,
  fonttitle=\sffamily\small, coltitle=white,
  boxrule=0.7pt, arc=2pt,
  left=2pt, right=2pt, top=2pt, bottom=2pt,
  toptitle=2pt, bottomtitle=2pt,
]
{\lstset{backgroundcolor=\color{bugbg}}\begin{lstlisting}
% Parameters (embed all data values directly here)
int: W = 6;
int: P = 5;

set of int: Warehouses = 1..W;
set of int: Ports = 1..P;

array[Warehouses] of string: wname =
  ["Verona","Perugia","Rome",
   "Pescara","Taranto","Lamezia"];
array[Ports] of string: pname =
  ["Genoa","Venice","Ancona","Naples","Bari"];

array[Warehouses] of int: supply =
  [10, 12, 20, 24, 18, 40];
array[Ports] of int: demand =
  [20, 15, 25, 33, 21];

% Truck capacity: 2 containers per truck
int: cap = 2;

% Distance (km) per container (for cost calculation)
array[Warehouses, Ports] of int: dist =
  [| 290, 115, 355, 715, 810
   | 380, 340, 165, 380, 610
   | 505, 530, 285, 220, 450
   | 655, 450, 155, 240, 315
   | 1010, 840, 550, 305, 95
   | 1072, 1097, 747, 372, 333 |];

int: cost_per_km = 30;
\end{lstlisting}}
{\lstset{backgroundcolor=\color{hlred}}\begin{lstlisting}
% Variables
array[Warehouses, Ports] of var 0..40: x;
array[Warehouses] of var 0..40: used;
array[Warehouses] of var 0..40: shipped;
var 0..1000000: total_cost;
\end{lstlisting}}
{\lstset{backgroundcolor=\color{bugbg}}\begin{lstlisting}
% Constraints
constraint
  forall(i in Warehouses) (
    shipped[i] = sum(j in Ports)(x[i,j])
  );
\end{lstlisting}}
{\lstset{backgroundcolor=\color{hlred}}\begin{lstlisting}
constraint                          % BUG 1
  forall(i in Warehouses) (
    shipped[i] = supply[i] - used[i]
  );
\end{lstlisting}}
{\lstset{backgroundcolor=\color{bugbg}}\begin{lstlisting}
constraint
  forall(j in Ports) (
    sum(i in Warehouses)(x[i,j]) = demand[j]
  );
\end{lstlisting}}
{\lstset{backgroundcolor=\color{hlred}}\begin{lstlisting}
% Each truck can carry up to 2      % BUG 2
% containers; count trucks
constraint
  forall(i in Warehouses) (
    used[i] = sum(j in Ports)
              (x[i,j]) div cap
  );
\end{lstlisting}}
{\lstset{backgroundcolor=\color{hlred}}\begin{lstlisting}
% Total cost                        % BUG 3
constraint
  total_cost = cost_per_km
    * sum(i in Warehouses)(used[i])
    * cap;
\end{lstlisting}}
{\lstset{backgroundcolor=\color{bugbg}}\begin{lstlisting}
% Objective
solve minimize total_cost;
\end{lstlisting}}
\end{tcolorbox}
\label{fig:minizinc-bugs}
\end{minipage}%
\hfill
\begin{minipage}[t]{0.485\textwidth}
\begin{tcolorbox}[
  title={\small\bfseries\sffamily (b) Corrected Code \\ Optimal cost: 904590},
  colback=fixbg, colframe=fixframe,
  fonttitle=\sffamily\small, coltitle=white,
  boxrule=0.7pt, arc=2pt,
  left=2pt, right=2pt, top=2pt, bottom=2pt,
  toptitle=2pt, bottomtitle=2pt,
]
{\lstset{backgroundcolor=\color{fixbg}}\begin{lstlisting}
% Parameters (embed all data values directly here)
int: W = 6;
int: P = 5;
set of int: Warehouses = 1..W;
set of int: Ports = 1..P;
array[Warehouses] of string: wname =
  ["Verona","Perugia","Rome",
   "Pescara","Taranto","Lamezia"];
array[Ports] of string: pname =
  ["Genoa","Venice","Ancona","Naples","Bari"];
array[Warehouses] of int: supply =
  [10, 12, 20, 24, 18, 40];
array[Ports] of int: demand =
  [20, 15, 25, 33, 21];
% Truck capacity: 2 containers per truck
int: cap = 2;
% Distance (km)
array[Warehouses, Ports] of int: dist =
  [| 290, 115, 355, 715, 810
   | 380, 340, 165, 380, 610
   | 505, 530, 285, 220, 450
   | 655, 450, 155, 240, 315
   | 1010, 840, 550, 305, 95
   | 1072, 1097, 747, 372, 333 |];
\end{lstlisting}}
{\lstset{backgroundcolor=\color{hlgreen}}\begin{lstlisting}
% Cost = 30 euros per km per CONTAINER
int: cost_per_km = 30;
% The cost to transport each container on
% route (i,j) is cost_per_km * dist[i,j]
array[Warehouses, Ports] of int: cost =
  array2d(Warehouses, Ports,
  [cost_per_km * dist[i,j]
   | i in Warehouses, j in Ports]);
% Variables
array[Warehouses, Ports] of var 0..40: x;
var 0..10000000: total_cost;
\end{lstlisting}}
{\lstset{backgroundcolor=\color{hlgreen}}\begin{lstlisting}
% Constraints
% Supply constraint: cannot ship more
% than available
constraint                          % FIX 1
  forall(i in Warehouses) (
    sum(j in Ports)(x[i,j]) <= supply[i]
  );
\end{lstlisting}}
{\lstset{backgroundcolor=\color{fixbg}}\begin{lstlisting}
% Demand constraint: each port must
% receive exactly its demand
constraint
  forall(j in Ports) (
    sum(i in Warehouses)(x[i,j]) = demand[j]
  );
\end{lstlisting}}
{\lstset{backgroundcolor=\color{hlgreen}}\begin{lstlisting}
% Total cost: 30 euros * distance   % FIX 2,3
% * number of containers on each route
constraint
  total_cost = sum(i in Warehouses,
                   j in Ports)
               (cost[i,j] * x[i,j]);
\end{lstlisting}}
{\lstset{backgroundcolor=\color{fixbg}}\begin{lstlisting}
% Objective
solve minimize total_cost;
\end{lstlisting}}
\end{tcolorbox}
\label{fig:phantom-bugs}
\end{minipage}

\vspace{6pt}

\begin{tcolorbox}[
  colback=codegray, colframe=black!40,
  boxrule=0.4pt, arc=2pt,
  left=6pt, right=6pt, top=4pt, bottom=4pt,
  title={\small\bfseries\sffamily Key Errors},
  fonttitle=\sffamily\small,
  toptitle=2pt, bottomtitle=2pt,
]
{\small
\textbf{Contradictory constraints (causes UNSATISFIABLE).}
Bugs 1 and 2 jointly force \texttt{used[i]} to satisfy two incompatible definitions:
\texttt{used[i] = supply[i] $-$ shipped[i]} and \texttt{used[i] = shipped[i] div 2}.
Substituting, this requires \texttt{supply[i] $-$ shipped[i] = shipped[i] div 2} for every warehouse,
which has no valid integer solution for most input values. The solver correctly reports the model as infeasible.
The fix removes \texttt{used[i]} and \texttt{shipped[i]} entirely and replaces them with a single
inequality \texttt{sum(x[i,j]) <= supply[i]}.

\vspace{4pt}
\textbf{Wrong objective formula.}
Bug 3 computes cost as \texttt{cost\_per\_km * sum(used[i]) * cap}, which
(a) never references the \texttt{dist} matrix, treating all routes as equally expensive, and
(b) models cost per truck rather than per container. The problem states the rate is 30 euros/km
\emph{per container}, so the correct objective is simply
\texttt{sum(cost\_per\_km * dist[i,j] * x[i,j])} over all pairs, with no dependence on truck capacity.
}
\end{tcolorbox}
\newpage
\subsection{Type II: Phantom Variables}
\label{app:phantom-example}
\textbf{\large LLM-Generated Phantom Variable and Index Errors in MiniZinc}
\vspace{4pt}
 
{\small\textbf{Problem Description.} A factory production planning LP (3 products, 3 equipment types, maximize profit). Full statement at: \url{https://huggingface.co/datasets/CardinalOperations/IndustryOr/viewer/default/test?q=3+products&row=53}}
 
\vspace{8pt}
 
\begin{minipage}[t]{0.485\textwidth}
\begin{tcolorbox}[
  title={\small\bfseries\sffamily (a) LLM-Generated Code --- Result: obj = 0},
  colback=bugbg, colframe=bugframe,
  fonttitle=\sffamily\small, coltitle=white,
  boxrule=0.7pt, arc=2pt,
  left=2pt, right=2pt, top=2pt, bottom=2pt,
  toptitle=2pt, bottomtitle=2pt,
]
{\lstset{backgroundcolor=\color{bugbg}}\begin{lstlisting}
% Parameters (embed all data values directly here)
int: N = 3;
set of int: P = 1..N;
int: M = 3;
set of int: E = 1..M;
 
% Equipment codes: 1=A, 2=B, 3=C
array[E] of int: avail = [300, 400, 420];
 
% Profit per unit (per thousand yuan):
% [3, 2, 2.9]
array[P] of float: profit =
  [3.0, 2.0, 2.9];
\end{lstlisting}}
{\lstset{backgroundcolor=\color{hlred}}\begin{lstlisting}
% Usage coefficients per unit:       % BUG 1
% [8, 2, 10] for I, II, III on A,B,C
array[P, E] of int: use =
  array2d(P, E,
[
  8, 2, 10,   % I on A,B,C
  10, 5, 8,   % II on A,B,C
  2, 13, 10   % III on A,B,C
]);
\end{lstlisting}}
{\lstset{backgroundcolor=\color{bugbg}}\begin{lstlisting}
% Variables
% x[p] = production quantity of product p
% (in "thousand yuan" units)
array[P] of var 0..100000: x;
\end{lstlisting}}
{\lstset{backgroundcolor=\color{hlred}}\begin{lstlisting}
% y[e,p] = whether to process       % BUG 2
% product p on equipment e (0/1)
array[E, P] of var 0..1: y;
 
% Constraints
% 1. Link decision variables to whether
% each product is processed on each
% equipment
constraint
  forall(e in E, p in P) (
    y[e,p] = sum(p in P)(x[p] * use[p,e])
  );
\end{lstlisting}}
{\lstset{backgroundcolor=\color{bugbg}}\begin{lstlisting}
% 2. Equipment capacity constraints
constraint
  forall(e in E) (
    sum(p in P)(x[p] * use[p,e]) <= avail[e]
  );
 
% Objective
solve maximize
  sum(p in P)(x[p] * profit[p]);
\end{lstlisting}}
\end{tcolorbox}
\end{minipage}%
\hfill
\begin{minipage}[t]{0.485\textwidth}
\begin{tcolorbox}[
  title={\small\bfseries\sffamily (b) Corrected Code --- Optimal obj = 135.26},
  colback=fixbg, colframe=fixframe,
  fonttitle=\sffamily\small, coltitle=white,
  boxrule=0.7pt, arc=2pt,
  left=2pt, right=2pt, top=2pt, bottom=2pt,
  toptitle=2pt, bottomtitle=2pt,
]
{\lstset{backgroundcolor=\color{fixbg}}\begin{lstlisting}
% Parameters (embed all data values directly here)
int: N = 3;
set of int: P = 1..N;
int: M = 3;
set of int: E = 1..M;
 
% Equipment codes: 1=A, 2=B, 3=C
array[E] of int: avail = [300, 400, 420];
 
% Profit per unit (per thousand yuan):
% [3, 2, 2.9]
array[P] of float: profit =
  [3.0, 2.0, 2.9];
\end{lstlisting}}
{\lstset{backgroundcolor=\color{hlgreen}}\begin{lstlisting}
% Usage coefficients:                % FIX 1
% rows = equipment, cols = products
array[E, P] of int: use =
  array2d(E, P,
[
  8, 2, 10,   % A on I, II, III
  10, 5, 8,   % B on I, II, III
  2, 13, 10   % C on I, II, III
]);
\end{lstlisting}}
{\lstset{backgroundcolor=\color{hlgreen}}\begin{lstlisting}
% Variables                          % FIX 2
% x[p] = production quantity of product p
array[P] of var 0.0..100000.0: x;
 
% (y variables removed entirely)
 
% Constraints
% Equipment capacity constraints
constraint
  forall(e in E) (
    sum(p in P)(x[p]
      * int2float(use[e,p]))
      <= int2float(avail[e])
  );
\end{lstlisting}}
{\lstset{backgroundcolor=\color{fixbg}}\begin{lstlisting}
% Objective
var float: obj =
  sum(p in P)(x[p] * profit[p]);
solve maximize obj;
\end{lstlisting}}
\end{tcolorbox}
\end{minipage}
 
\vspace{6pt}
 
\begin{tcolorbox}[
  colback=codegray, colframe=black!40,
  boxrule=0.4pt, arc=2pt,
  left=6pt, right=6pt, top=4pt, bottom=4pt,
  title={\small\bfseries\sffamily Key Errors},
  fonttitle=\sffamily\small,
  toptitle=2pt, bottomtitle=2pt,
]
{\small
\textbf{Phantom binary variables force objective to zero.}
Bug 2 introduces binary variables \texttt{y[e,p]} (bounded 0..1) and constrains each to equal
\texttt{sum(p in P)(x[p] * use[p,e])}, a quantity that can reach hundreds.
The only way to satisfy \texttt{0 <= result <= 1} is to set all \texttt{x[p] = 0},
which the solver does, producing an objective of zero.
These variables have no role in the problem (it is a standard LP, not an assignment problem)
and are removed entirely in the fix.
 
\vspace{4pt}
\textbf{Transposed index on the \texttt{use} matrix.}
Bug 1 declares the matrix as \texttt{array[P,\,E]} (products $\times$ equipment) but fills it
row-by-row from the problem table, whose rows are equipment.
The first data row \texttt{[8, 2, 10]} represents equipment A's hours for products I, II, III,
not product I's hours on A, B, C.
With the wrong index the capacity constraint reads the wrong cell, so even if the phantom-variable
bug were absent the solution would still be incorrect.
The fix changes the declaration to \texttt{array[E,\,P]} and accesses it as \texttt{use[e,p]}.
}
\end{tcolorbox}

\begin{table}[htbp]
\centering
\caption{Two categories of exclusive constraint errors observed in small fine-tuned LLMs (our augmented fine-tuning variant) and their impact on solution accuracy. \textit{Type I}: contradictory constraints causing the solver to return unsatisfiable. \textit{Type II}: phantom variables forcing the objective to zero. Fixing these two classes of errors in combination has a considerable potential (52\%) to improve solution accuracy, even improving over GPT-5.2@1 (49\%).}
\label{tab:constraint-errors}
\vspace{6pt}
\renewcommand{\arraystretch}{1.25}
 
\textbf{Type I: Contradictory Constraints (Unsatisfiable)}
\vspace{4pt}
 
\begin{tabular}{l c c c}
\toprule
\textbf{Model} & \textbf{Solution Acc. (\%)} & \textbf{Unsat Cases} & \textbf{Potential Improvement (\%)} \\
\midrule
qwen3-0.6b-augmented    & 13.0 & 18  & 31.0 \\
llama-3.2-1b-augmented  &  8.0 & 13  & 21.0 \\
llama-3.2-3b-augmented  & 17.0 & 15  & 32.0 \\
gemma-2-9b-augmented    & 22.0 & 17  & 39.0 \\
gpt-oss-20b-augmented   & 32.0 &  8  & 40.0 \\
\bottomrule
\end{tabular}
 
\vspace{14pt}
 
\textbf{Type II: Phantom Variables Forcing Objective to Zero}
\vspace{4pt}
 
\begin{tabular}{l c c c c}
\toprule
\textbf{Model} & \textbf{Solution Acc. (\%)} & \textbf{Zero-Obj Cases} & \textbf{Potential Improvement (\%)} \\
\midrule
qwen3-0.6b-augmented    & 13.0 & 12 & 25.0 \\
llama-3.2-1b-augmented  &  8.0 & 20 & 28.0 \\
llama-3.2-3b-augmented  & 17.0 &  7 & 24.0 \\
gemma-2-9b-augmented    & 22.0 & 18 & 40.0 \\
gpt-oss-20b-augmented   & 32.0 & 12 & 44.0 \\

\bottomrule
\end{tabular}
\end{table}

\end{document}